\documentclass[10pt,twocolumn,letterpaper]{article}

\usepackage[pagenumbers]{cvpr}

\usepackage{cuted} %

\usepackage{currfile} %

\usepackage{caption} %

\usepackage{amsmath,amssymb}
\usepackage{booktabs}
\usepackage{colortbl}  %
\definecolor{oursrow}{rgb}{0.826,0.888,0.943}  %
\usepackage{multirow}
\usepackage{xspace}
\usepackage{algorithm}
\usepackage[noend]{algpseudocode}
\usepackage{tikz}
\usetikzlibrary{positioning,arrows.meta,fit,calc}
\newcommand{\method}{HorizonFlow\xspace}
\newcommand{\sg}{\operatorname{sg}}
\newcommand{\xG}{x_{G}}
\newcommand{\xt}{\tilde{x}}
\newcommand{\dce}{\Delta_{\mathrm{CE}}}
\newcommand{\dsc}{\Delta_{\mathrm{SC}}}
\newcommand{\xreal}{\hat{x}_{\mathrm{real}}}
\newcommand{\xfake}{\hat{x}_{\mathrm{fake}}}
\newcommand{\Edata}{\bar{E}_{\mathrm{data}}}
\usepackage{iftex}
\ifXeTeX
  \usepackage{fontspec}
\fi

\definecolor{cvprblue}{rgb}{0.21,0.49,0.74}
\usepackage[pagebackref,breaklinks,colorlinks,allcolors=cvprblue]{hyperref}
\AtBeginDocument{\crefname{appendix}{Supp.\ Sec.}{Supp.\ Secs.}\Crefname{appendix}{Supplementary Section}{Supplementary Sections}}

\newcommand{\codeurl}{{\Urlmuskip=0mu\relax\url{https://github.com/XIANGLONGGUO/HorizonFlow-repo}}}
\AtEndEnvironment{abstract}{\par\vspace{0.5em}\noindent Code: {\small\href{https://github.com/XIANGLONGGUO/HorizonFlow-repo}{\texttt{github.com/XIANGLONGGUO/HorizonFlow-repo}}}}
\newcommand{\suppcodesentence}{The code is available at \codeurl.}

\title{HorizonFlow: Few-Step Video Distillation in 600 Iterations}

\author{
Xianglong Guo$^{1}$ \qquad Haobo Xu$^{2}$ \qquad Zanyi Wang$^{3}$ \qquad Zhen Lei$^{4}$ \qquad Chang Yu$^{1,\dagger}$\\[0.3em]
$^{1}$Nanjing University \qquad $^{2}$Tsinghua University \qquad $^{3}$University of California, San Diego\\
$^{4}$Institute of Automation, Chinese Academy of Sciences\\[0.3em]
{\small $^{\dagger}$Corresponding author}
}

\begin{document}
\makeatletter\def\@crefstar#1#2{\@cref{#1}{#2}}\makeatother

\maketitle
\begin{strip}
\centering
\includegraphics[width=\linewidth]{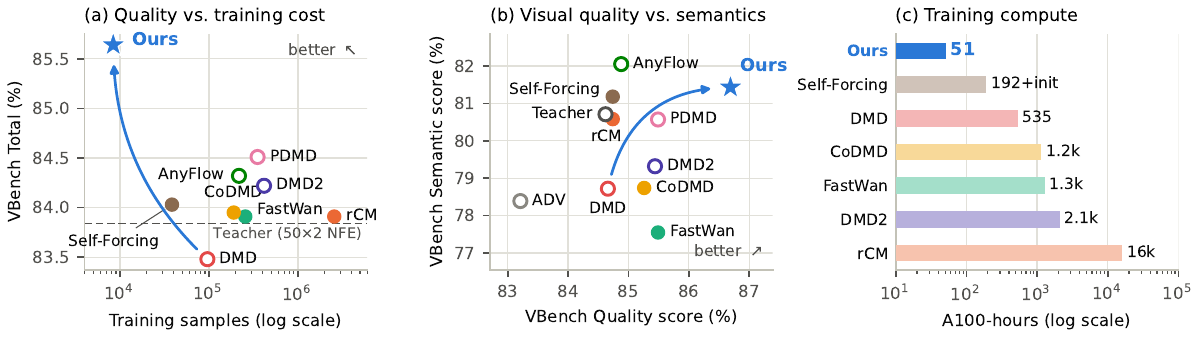}
\captionsetup{hypcap=false}
\captionof{figure}{\textbf{Few-step video distillation in 600 iterations.}
\textbf{(a)} \method reaches a VBench total score of 85.64 with 8.4k training samples, 11$\times$ fewer than DMD and 304$\times$ fewer than rCM.
\textbf{(b)} It has the highest visual quality score of all models.
\textbf{(c)} It trains with 3.8--314$\times$ fewer A100-hours.
Filled markers are measured by us, hollow markers are reported in~\cite{pdmd}; details are in \cref*{sec:supp_eval}.}
\captionsetup{hypcap=true}
\label{fig:teaser}%
\vspace{-1em}
\end{strip}

\begin{abstract}
Distribution matching distillation (DMD) enables fast video generation by distilling a multi-step teacher into a few-step student. However, DMD-based methods often suffer from unnatural motion and costly training.
We trace these limitations to the DMD gradient, which jointly optimizes appearance, layout, and motion without distinguishing their learning dynamics.
Its static component, which carries appearance and layout, has a much higher signal-to-noise ratio than its motion component. As a result, per-frame quality converges within a few hundred iterations, whereas motion lags far behind.
Beyond being noisy, the gradient also lacks two signals.
(1) Video VAEs encode several frames into one latent frame (four in Wan2.1), and the gradient provides little guidance for spreading motion over these frames, causing periodic stutter.
(2) DMD does not constrain the final sampling step to preserve the layout set by earlier steps, so objects merge and drift once the video moves.
We propose \method, which decouples the correction of motion from that of layout. It keeps the DMD loss and only shifts its regression target, with one training-only correction in the motion component and one in the orthogonal layout component.
Chunk Evenness addresses (1) by spreading the motion of each latent chunk evenly over its frames, and Semantic Commitment addresses (2) by anchoring the coarse layout to the teacher's completion of the student's last step.
In only 600 iterations, \method distills Wan2.1-T2V-1.3B into a 4-step model using 8.4k training samples, 4.6--304$\times$ fewer than prior work.
On VBench, \method outperforms all released few-step Wan2.1-T2V-1.3B models. The same recipe transfers to Cosmos-Predict2.5-2B, where our student outperforms NVIDIA's distilled model.
\end{abstract}

\section{Introduction}
\label{sec:intro}

\begin{figure*}[t]
\centering
\includegraphics[width=\textwidth]{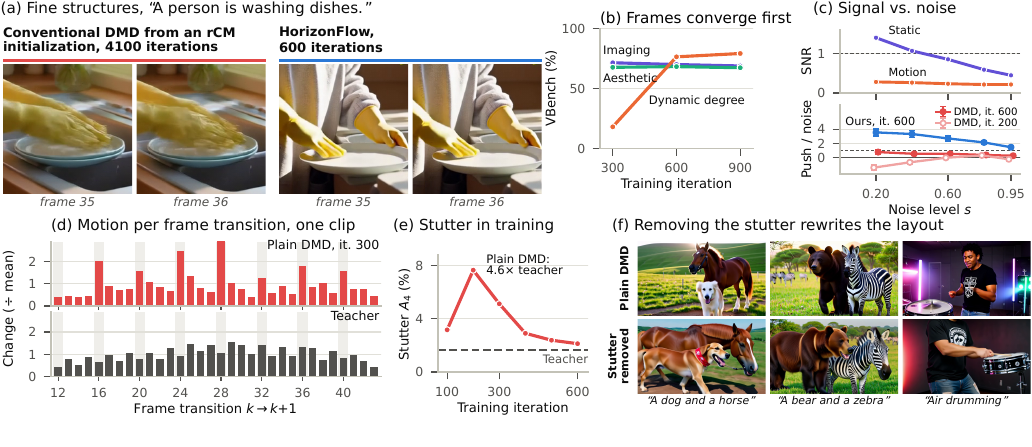}
\caption{\textbf{Video DMD learns appearance before motion, and correcting the motion disturbs the layout.}
(a) After 4100 iterations of conventional DMD training, the glove's fingers lose their shape from frame to frame; \method keeps them intact after 600.
(b) Plain DMD: imaging and aesthetic quality reach their final level by 300 iterations, while the dynamic degree is still 18.
(c) Top: signal-to-noise ratio over noise draws of the static and motion components of the plain DMD gradient (dashed: signal = noise).
Bottom: push of the update toward evenly spread motion, divided by its noise; the DMD gradient stays below the noise (dashed), while our update (DMD + CE + SC, same samples) exceeds it at every noise level.
(d) Motion per frame transition in one clip, divided by its mean: plain DMD (300 iterations) moves mostly at the transitions between latent chunks (shaded); the teacher moves evenly.
(e) The stutter amplitude $A_4$, the period-4 component of these changes, peaks at 4.6 times the teacher's level at 200 iterations.
(f) Chunk Evenness alone (600 iterations) removes the stutter and raises the dynamic degree from 76 to 97, but the dog fuses with the horse, the bear with the zebra, and the drummer's head leaves the frame.
Details are in \cref*{sec:supp_metrics,sec:supp_grad}.}
\label{fig:obs}
\vspace{-1em}
\end{figure*}

Video diffusion models such as Wan2.1~\cite{wan2025} and Cosmos-Predict2.5~\cite{cosmos25} generate high-quality videos from text.
However, they need dozens of denoising steps, and classifier-free guidance~\cite{cfg} doubles the cost of each step.
Distribution matching distillation (DMD)~\cite{dmd,dmd2} trains a student to match the teacher's output distribution, enabling few-step generation, and is widely used for fast video generation~\cite{causvid,selfforcing,rcm,vsa,codmd,pdmd}.
Current few-step students render sharp frames, but remain expensive to train and often exhibit unnatural motion (\cref{fig:obs}a).

Prior work treats video DMD mainly as a stability problem.
It attributes artifacts and collapse to errors of the critic, the auxiliary network that estimates the score of the student's outputs~\cite{pdmd,codmd}.
Its remedies therefore make the critic more accurate or the student update more cautious, for example, with several critic updates per student update~\cite{dmd2}, small learning rates~\cite{vsa,pdmd}, or a separate initialization stage~\cite{rcm,causvid,selfforcing,anyflow}.
These methods are costly, using 38k to 2.56M training samples to distill Wan2.1-1.3B (\cref{fig:teaser}a).
However, high training cost does not guarantee smooth motion, since these remedies do not add the signals the DMD gradient lacks: rCM~\cite{rcm}, trained on 2.56M samples, has the least smooth motion of all models we evaluate.

We take a different view and examine the learning dynamics of DMD itself, whose gradient updates appearance, layout, and motion with one entangled signal.
In a minimal setup without these remedies, DMD learns \emph{static} visual properties much faster than \emph{motion}: imaging and aesthetic quality on VBench converge within 300 iterations, while the dynamic degree is still far from its final level (\cref{fig:obs}b).
Motion is also uneven.
Wan2.1 encodes every four frames into one latent frame, and the student packs most of a chunk's motion into one or two of its four frame transitions, whereas the teacher, with the same VAE, moves evenly.
The video therefore stutters every fourth frame (\cref{fig:obs}d and e), as do the released models trained with other recipes (\cref*{tab:supp_severe}).
Both observations follow from the DMD gradient.
We split it into a static component, averaged across frames, and a motion component, the remaining temporal variation.
Across noise draws, the static component has a much higher signal-to-noise ratio than the motion component, which is dominated by noise (\cref{fig:obs}c, top), and the push toward evenly distributed motion stays below its noise (\cref{fig:obs}c, bottom).
Correcting motion alone, however, is not enough.
The stutter disappears, but objects merge and layouts drift (\cref{fig:obs}f), because a few-step sampler fixes the layout in its early steps and DMD never asks the last step to keep it.
The DMD gradient thus lacks two signals: even motion within a latent chunk, and a last step that preserves the layout.

We therefore propose \method (\cref{fig:overview}), which decouples motion from layout in the update: it leaves the DMD gradient unchanged and supplies each missing signal with its own correction of the training target, one in the motion component and one in the layout component, which is orthogonal to motion.
(1)~\emph{Chunk Evenness} (CE) decodes a short window of the student's video and, where the motion within a chunk is less even than in the training clips, shifts the target to spread this motion over all four frames.
(2)~\emph{Semantic Commitment} (SC) lets the teacher complete the student's last step from the same input and pulls only the coarse, time-averaged layout of the output toward this completion, which holds objects in place without opposing their movement.
The two corrections add a short-window decode and one teacher prediction per sample, about 9\% of the training time, and nothing to inference.
Unlike the DMD gradient, our update pushes toward even motion by more than its noise (\cref{fig:obs}c, bottom), so the student learns even motion and layout within 600 iterations.
On Wan2.1-T2V-1.3B, \method trains for only 600 iterations (51 A100-hours), with 4.6--304$\times$ fewer training samples and 3.8--314$\times$ fewer A100-hours than prior distillations (\cref{fig:teaser}).
It raises the VBench total score by 1.61 points over the best released few-step model and by 1.13 over plain DMD at nearly the same training cost, the latter mostly through motion.
The same recipe transfers to Cosmos-Predict2.5-2B~\cite{cosmos25}, where our student scores 0.66 points above NVIDIA's distilled model, which additionally uses real data and an adversarial loss.

Our contributions are:
\begin{itemize}
\item By analyzing the DMD gradient in video distillation, we find that its motion component is dominated by noise and that it lacks two signals, even motion within a latent chunk and a last step that keeps the layout.
\item We propose \method, which decouples the update by adding the two missing signals as separate corrections of the training target, Chunk Evenness for motion and Semantic Commitment for layout, at no inference cost.
\item \method gives the best few-step Wan2.1-T2V-1.3B model on VBench at a fraction of the training cost of prior work, and transfers to Cosmos-Predict2.5-2B.
\end{itemize}

\section{Related Work}
\label{sec:related}

\paragraph{Few-step diffusion distillation.}
Trajectory distillation compresses the teacher's sampling path: progressive distillation halves the number of steps repeatedly~\cite{progressive}, consistency models map every point of the path to its end~\cite{cm,lcm,scm}, and flow-map and transition models learn jumps between noise levels~\cite{anyflow,tmd}.
Distribution distillation instead matches the distribution of the student's outputs to the teacher's, with a discriminator~\cite{add,apt,dmad} or with score differences~\cite{vsd,dmd,dmd2,sid}.
DMD~\cite{dmd,dmd2} updates the student with the difference between the teacher's score and that of a critic fitted online to the student's samples.
Without DMD2's GAN term, it needs neither teacher trajectories nor real data, and it has become the default for video.

\paragraph{Making DMD work for video.}
Most video methods keep the DMD objective and stabilize it.
DMD2~\cite{dmd2} runs several critic updates per student update, following the two time-scale update rule~\cite{ttur}.
Self-Forcing~\cite{selfforcing}, rCM~\cite{rcm}, and CoDMD~\cite{codmd} use it with learning rates of $2{\times}10^{-6}$, and FastWan~\cite{vsa} trains on 600k teacher-generated videos.
PDMD~\cite{pdmd} projects estimated critic errors out of the update and lowers the learning rate to $4{\times}10^{-7}$, and DyMD~\cite{dymd} adapts critic fitting and noise levels to strong-motion rollouts.
A second family precedes DMD with an initialization stage: consistency distillation~\cite{scm} in rCM~\cite{rcm}, regression onto teacher trajectories in CausVid and Self-Forcing~\cite{causvid,selfforcing}, and flow maps in AnyFlow~\cite{anyflow}.
A third family changes the objective or adds video-specific regularizers: phase-wise experts~\cite{phaseddmd}, relations across frames and samples~\cite{codmd}, temporal regularization against oversaturation and collapse~\cite{adv}, and restored diversity in autoregressive rollouts~\cite{udmd}.
SGMD~\cite{sgmd} changes the gradient form and recovers motion but reports a loss in semantics, as does CE alone (\cref{sec:ablation}).
Instead of stabilizing the critic, \method decomposes the DMD gradient and adds the two signals that it lacks and no method above provides, each in its own component: even motion inside a VAE chunk and a last sampling step that keeps the committed layout.

\paragraph{Latent video models.}
Modern video generators run diffusion in the latent space of a VAE~\cite{ldm,vdm} with a transformer backbone~\cite{dit}.
Their VAEs compress time causally, typically four frames into one latent frame~\cite{wan2025,cogvideox,hunyuanvideo,cosmos25}.
This compression makes training and sampling affordable, but the latent-space DMD gradient carries little signal against the chunk-level stutter (\cref{sec:analysis}).
CE supplies this signal at training time without changing the VAE or the generator.

\section{Preliminaries}
\label{sec:prelim}

DMD uses three components: a frozen teacher, a few-step student initialized from it, and a critic.
Training alternates two updates: the critic learns to denoise student videos, and the student moves each video toward the teacher's prediction of its noisy version, away from the critic's.

\paragraph{Teacher and student.}
The teacher is a rectified-flow model~\cite{flowmatching,rectifiedflow}.
It turns a clean latent video $x_0$ into Gaussian noise $\epsilon$ along a straight path and predicts the velocity along this path.
Following~\cite{scm,rcm}, we index noise levels by the TrigFlow time $t\in[0,\pi/2]$, with $x_t=\cos t\,x_0+\sin t\,\epsilon$.
When we state noise thresholds, we use the equivalent rectified-flow time $s=\sin t/(\cos t+\sin t)\in[0,1]$.
From its velocity, a network $f$ gives a clean-video prediction $\hat x_f(x_t,t,c)$ for a prompt $c$.
Classifier-free guidance~\cite{cfg} with scale $w$ extrapolates from the unconditional toward the conditional prediction and gives the guided prediction $\hat x^{w}_f$.

At inference, the student $G_\theta$ generates a video in four steps: each step predicts the clean video and re-noises it to a lower noise level, and the last prediction is the output $\xG$.
In training, it rolls out $k$ steps from pure noise, with $k$ cycling through $1,\dots,4$ and random intermediate times, and gradients flow through the last step only~\cite{selfforcing}.

\paragraph{Distribution matching distillation.}
DMD~\cite{dmd,dmd2} re-noises $\xG$ at a random time $t$ and compares the guided teacher prediction $\xreal=\hat x^{w}_\phi(x_t,t,c)$ with that of the critic $\psi$, $\xfake=\hat x_\psi(x_t,t,c)$.
Their normalized difference is the DMD gradient $g$, and the student regresses onto the pseudo-target $\xt=\sg(\xG-g)$, where $\sg$ stops gradients:
\begin{align}
g&=\frac{\xfake-\xreal}{\omega},\qquad \omega=\sg\Big(\tfrac{1}{N}\big\lVert \xG-\xreal\big\rVert_1\Big),
\label{eq:dmd}\\
\mathcal{L}_G&=\tfrac12\big\lVert \xG-\xt\big\rVert_2^2,
\label{eq:target}
\end{align}
with $N$ the number of latent entries.

\paragraph{Latent chunks.}
The Wan2.1 VAE compresses time causally: the first frame becomes the first latent frame, and every following group of four frames, a \emph{chunk}, becomes one latent frame.
An 81-frame clip thus has 21 latent frames, latent frame $l\geq1$ decodes to frames $4l-3,\dots,4l$, and we call the position of a frame inside its chunk its \emph{phase}.

\section{Method}
\label{sec:method}

\begin{figure*}[t]
\centering
\includegraphics[width=\textwidth]{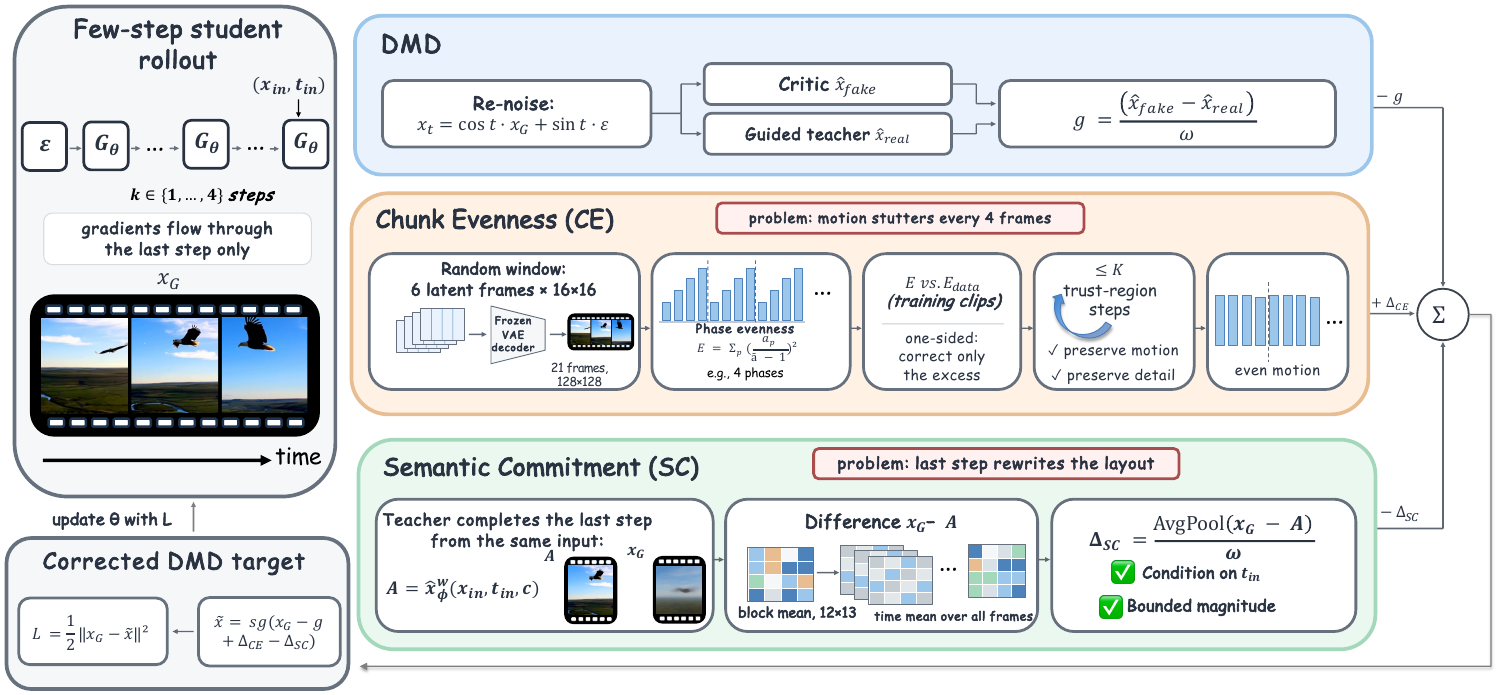}
\caption{\textbf{One student update of \method.}
The student rolls out $k\in\{1,\dots,4\}$ steps from noise; gradients flow only through the last step, which maps $(x_{\mathrm{in}},t_{\mathrm{in}})$ to $\xG$.
\emph{DMD} (blue) re-noises $\xG$ and compares the critic with the guided teacher, giving the DMD gradient $g$.
\emph{Chunk Evenness} (orange) decodes a random window of $\xG$ with the frozen VAE, measures how evenly its motion is spread over the four frame phases of each latent chunk, and, where it is less even than the training clips, takes a few normalized steps that keep motion and detail, giving $\dce$.
\emph{Semantic Commitment} (green) lets the teacher complete the student's last step from the same input and keeps only the block- and time-averaged difference, gated by $t_{\mathrm{in}}$ and capped, as $\dsc$.
Both corrections only move the DMD target, $\xt=\sg(\xG-g+\dce-\dsc)$, each in its own component of the update (CE: motion; SC: layout); the critic update, the sampler, and inference are those of a plain 4-step student.}
\label{fig:overview}

\vspace{-1em}
\end{figure*}

\method keeps the DMD objective of \cref{sec:prelim} and changes only its pseudo-target $\xt$ (\cref{fig:overview}).
\Cref{sec:obs} describes the training setup and splits the DMD gradient into components.
\Cref{sec:ce,sec:sc} then add one correction for motion and one for layout, and \cref{sec:full} combines them into a single update.
\Cref*{tab:supp_hparams} lists all hyperparameters.

\subsection{Setup and gradient decomposition}
\label{sec:obs}

We train plain DMD (\cref{eq:dmd,eq:target}), \ie DMD without our corrections, in a minimal setup: the student is initialized from the teacher without a separate initialization stage, the critic is updated once per student update, the teacher uses its default guidance scale, and both networks use standard fine-tuning learning rates.
We decompose its gradient $g$ with two projections: $M$ replaces every latent frame by the mean over all latent frames, and $\bar P$ further averages each frame over a coarse grid of blocks.
Both are orthogonal projections, so
\begin{equation}
g=\underbrace{\bar P g}_{\text{layout}}+\underbrace{(M-\bar P)\,g}_{\text{detail}}+\underbrace{(I-M)\,g}_{\text{motion}}
\label{eq:decomp}
\end{equation}
splits $g$ into mutually orthogonal components.
The first two form the static component $Mg$, which every frame shares, and the third is the motion component.
In plain DMD, the static component carries a consistent signal, whereas the motion component is dominated by noise, and the gradient barely pushes toward even motion within a latent chunk (\cref{sec:analysis}).
Moreover, no component asks the last sampling step to keep the layout that the earlier steps have committed to.
CE and SC supply these two missing signals.

\subsection{Chunk Evenness}
\label{sec:ce}

Plain DMD packs each latent chunk's motion into one or two of its four frames, so the video stutters every fourth frame (\cref{fig:obs}d).
This stutter is a property of the decoded frames rather than of the latents.
Chunk Evenness (CE) therefore measures it in pixel space and corrects the student only where its motion is less even than in the training clips.
From each student output $\xG$, we cut a small window $z$ of consecutive latent frames at a random position and decode it with the frozen VAE into grayscale frames.
From the RMS differences $r_j$ of consecutive frames, we form the phase means $a_p=\operatorname{mean}_{j\equiv p\ (\mathrm{mod}\ 4)} r_j$, $p=0,\dots,3$, with average $\bar a$, and define the phase evenness
\begin{equation}
E(z)=\sum_{p=0}^{3}\Big(\frac{a_p}{\bar a}-1\Big)^2 .
\label{eq:evenness}
\end{equation}
$E$ is zero when motion is spread evenly over the four phases, and it is differentiable through the decoder.
Driving $E$ to zero would be wrong, because a frozen video is perfectly even.
CE therefore corrects only the unevenness in excess of the training clips.
It tracks their evenness $\Edata$ as a moving average on random windows of training latents and computes the excess with a margin $m$,
\begin{equation}
e=\operatorname{clip}\big(E(z)/\Edata-1-m,\;0,\;1\big).
\label{eq:excess}
\end{equation}
For $e>0$, CE moves the window with a few normalized gradient steps,
\begin{equation}
z\leftarrow z-\eta\,e\,\sigma_z\,\frac{\nabla_z E}{\operatorname{rms}(\nabla_z E)},
\label{eq:ce_step}
\end{equation}
where $\sigma_z$ is the RMS temporal difference of the window and $\eta$ is a small step size.
A step is kept only if it lowers $E$ and keeps both the decoded motion and the high-frequency energy of the frames close to their initial values.
This rules out the two cheap ways to look even, freezing and blurring.
The accepted displacement forms $\dce$, which is zero outside the window, and the pseudo-target becomes $\xt+\dce$.
CE thus never pushes evenness beyond the training clips and never removes motion or detail; it only redistributes the motion of each chunk over its four frames.

\subsection{Semantic Commitment}
\label{sec:sc}

A few-step sampler fixes the layout of the video in its early steps.
Semantic Commitment (SC) asks the last student step to keep this layout and leaves everything else to DMD and CE.
In a rollout with $k\geq2$ steps, let $(x_{\mathrm{in}},t_{\mathrm{in}})$ be the input of the last student step and $\xG$ its output.
We let the teacher take the same step from the same input,
\begin{equation}
A=\hat x^{w}_\phi(x_{\mathrm{in}},t_{\mathrm{in}},c).
\label{eq:anchor}
\end{equation}
The anchor $A$ starts from the student's own state, so it inherits the layout that the student committed to in its earlier steps.
Unlike regression onto teacher trajectories~\cite{causvid,selfforcing}, it needs no stored trajectories.

A one-step teacher prediction is blurry in its details but reliable in its layout.
SC therefore compares $\xG$ and $A$ only in the layout component of \cref{eq:decomp}:
\begin{equation}
\dsc=\begin{cases}\lambda\,\dfrac{\bar P(\xG-A)}{\omega} & \text{if } s(t_{\mathrm{in}})<\tau,\\[4pt] 0 & \text{otherwise,}\end{cases}
\label{eq:sc}
\end{equation}
with a weight $\lambda$, a threshold $\tau$ on the rectified-flow noise level $s$, and $\omega$ the DMD normalizer of \cref{eq:dmd}; the pseudo-target becomes $\xt-\dsc$.
The block mean discards detail, so SC does not copy the blurry texture of the teacher.
The time mean makes $\dsc$ orthogonal to the motion component, so SC fixes which objects are present and where they are, but never opposes how they move.
The condition on $s(t_{\mathrm{in}})$ skips last inputs that are still almost pure noise, where no layout has been committed yet.
Finally, a per-sample cap on $\lVert\dsc\rVert_2$ relative to $\lVert g\rVert_2$ keeps SC from dominating the DMD update.

\subsection{The decoupled update}
\label{sec:full}

\begin{table*}[t]
\centering
\caption{\textbf{VBench on Wan2.1-T2V-1.3B.} 946 prompts, 5 videos each (seeds 42--46; teacher: 1). Each score is the mean over the five seeds $\pm$ its standard deviation. Samples = optimizer iterations of student and critic $\times$ global batch, as rCM, FastWan, and Self-Forcing count them, from each paper. A100-h: training compute, measured for \method, reported for CoDMD, and estimated from the published sample budget for the others (\cref*{sec:supp_eval}).}
\label{tab:main}
\small
\setlength{\tabcolsep}{6.5pt}
\setlength{\aboverulesep}{0pt}\setlength{\belowrulesep}{0pt}\renewcommand{\arraystretch}{1.2}  %
\begin{tabular}{lrrcccccc}
\toprule
Method & Samples & A100-h & Total & Quality & Semantic & Multi-obj. & Spatial \\
\midrule
Teacher (50 steps) & -- & -- & 82.81 & 83.56 & 79.81 & 70.50 & 79.62 \\
\midrule
FastWan~\cite{vsa} (3 steps) & 256k & 1.3k & 83.91{\scriptsize$\pm$0.30} & 85.49{\scriptsize$\pm$0.40} & 77.55{\scriptsize$\pm$0.25} & 75.24{\scriptsize$\pm$2.21} & 73.83{\scriptsize$\pm$2.90} \\
CoDMD~\cite{codmd} & $\sim$190k & 1.2k & 83.95{\scriptsize$\pm$0.22} & 85.26{\scriptsize$\pm$0.24} & 78.74{\scriptsize$\pm$0.49} & 80.78{\scriptsize$\pm$1.70} & 73.19{\scriptsize$\pm$3.54} \\
rCM~\cite{rcm} & 2.56M & 16k & 83.91{\scriptsize$\pm$0.22} & 84.74{\scriptsize$\pm$0.29} & 80.58{\scriptsize$\pm$0.43} & 84.54{\scriptsize$\pm$1.76} & 81.26{\scriptsize$\pm$3.35} \\
Self-Forcing~\cite{selfforcing} (autoregressive) & 38.4k + init. & 192 + init. & 84.03{\scriptsize$\pm$0.16} & 84.74{\scriptsize$\pm$0.15} & 81.18{\scriptsize$\pm$0.24} & 80.85{\scriptsize$\pm$4.43} & 82.91{\scriptsize$\pm$1.74} \\
\midrule
\rowcolor{oursrow}
\method & 8.4k & 51 & \textbf{85.64}{\scriptsize$\pm$0.16} & \textbf{86.69}{\scriptsize$\pm$0.18} & \textbf{81.43}{\scriptsize$\pm$0.40} & \textbf{85.27}{\scriptsize$\pm$1.11} & \textbf{88.32}{\scriptsize$\pm$2.09} \\
\bottomrule
\end{tabular}
\vspace{-1em}
\end{table*}

\method trains the student with the DMD loss of \cref{eq:target} on the corrected target
\begin{equation}
\xt=\sg\big(\xG-g+\dce-\dsc\big)
\label{eq:full}
\end{equation}
(\cref*{alg:update}).
Each signal acts in its own component of \cref{eq:decomp}.
DMD keeps its consistent signal in the static component.
CE redistributes motion within chunks and thus acts mainly on the motion component, while SC acts on the layout component, which is orthogonal to motion.
Both corrections are computed without gradients to the student and only move the target.
They add one window decode with up to four decoder passes for CE and one guided teacher prediction for SC.
The critic update, the sampler, and inference are unchanged, so the distilled model generates exactly like a plain 4-step student.

\section{Experiments}
\label{sec:exp}

We compare \method with all released few-step models of Wan2.1-T2V-1.3B (\cref{sec:main}), measure why plain DMD lags in motion (\cref{sec:analysis}), test \method with fewer sampling steps and on a second backbone (\cref{sec:transfer}), and analyze the role of each component (\cref{sec:ablation}).

\subsection{Setup}
\label{sec:setup}

\paragraph{Implementation details.}
We distill Wan2.1-T2V-1.3B~\cite{wan2025} at 832$\times$480 and 81 frames, with the synthetic Wan2.1-14B clips of rCM~\cite{rcm}: DMD uses their captions, CE their latents for $\Edata$.
All models follow the recipe of \cref{sec:obs} on seven A100 GPUs with a batch of 14 for 600 iterations.
An iteration of \method takes 43.6\,s, 9\% more than plain DMD; the run takes 51 A100-hours.
FastWan and CoDMD distill with guidance $w=3.5$ instead of Wan's default of 5, which changes plain DMD by only 0.04 points, so it does not explain our lead.
The student samples at TrigFlow times $(\arctan 1600, 1.5, 1.4, 1.0)$; \cref*{sec:supp_impl} lists all hyperparameters.

\paragraph{Baselines.}
We compare against the released checkpoints of four few-step distillations of Wan2.1-T2V-1.3B, each with its official sampler.
rCM~\cite{rcm} precedes DMD with consistency distillation, CoDMD~\cite{codmd} adds relational structure across frames and samples, FastWan~\cite{vsa} combines DMD with sparse attention and samples in three steps, and Self-Forcing~\cite{selfforcing} trains an autoregressive student on its own rollouts.
\method instead keeps the DMD objective and adds the two missing signals as corrections of its target.

\paragraph{Evaluation protocol.}
We evaluate with VBench~\cite{vbench} on 946 GPT-4o-augmented prompts, with five videos per prompt for the main comparisons and one for the ablations.
VBench scores 16 dimensions, seven of visual quality, such as motion smoothness and dynamic degree, and nine of semantics, such as multiple objects and spatial relationship, and combines them into quality, semantic, and total scores.

\subsection{Main results}
\label{sec:main}

\renewcommand{\dblfloatpagefraction}{0.8}
\begin{figure*}[p]
\centering
\includegraphics[width=\textwidth,height=0.79\textheight,keepaspectratio]{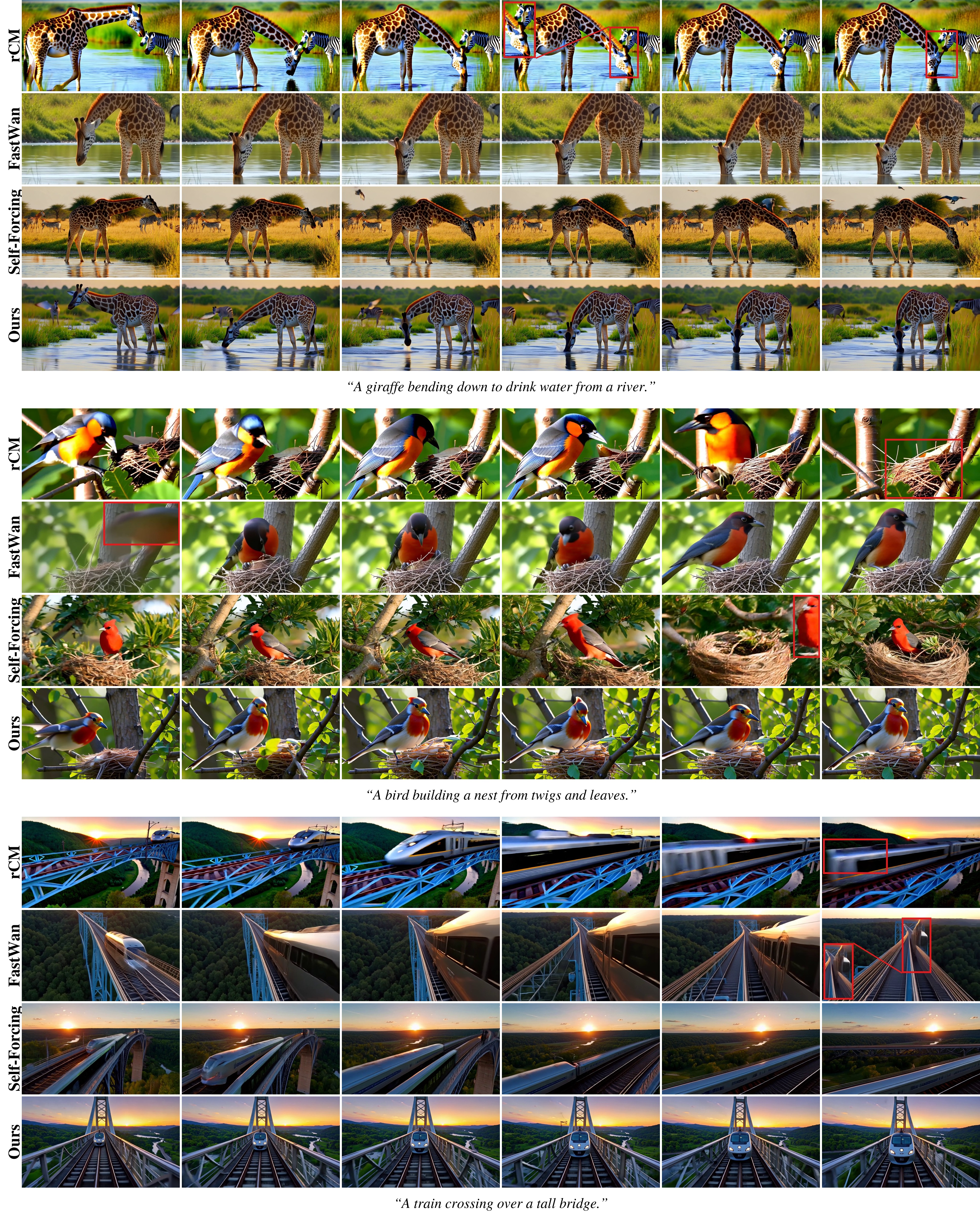}
\vspace{-4pt}
\caption{\textbf{Qualitative comparison with released few-step models on Wan2.1-T2V-1.3B.}
Every method gets the same VBench prompt and seed; columns are frames 1, 17, 33, 49, 65 and 81 of 81.
Red boxes mark failures; insets enlarge them.
rCM fuses the giraffe's head with a zebra, loses the bird from its nest and smears the train.
FastWan barely moves the giraffe, opens the bird clip with a blurred double exposure and melts the train into the rails.
Self-Forcing cuts to a new shot halfway through the bird clip and pushes the bird out of the frame.
\method carries out each prompt: the giraffe bends down and drinks, the bird settles on its nest, and the train crosses the bridge toward the camera, with every subject intact from the first frame to the last.
CoDMD and more prompts are in the supplementary material.}
\label{fig:qualitative}
\end{figure*}

\paragraph{Quantitative comparison.}
\Cref{tab:main} compares the released few-step models of Wan2.1-1.3B.
\method leads on every score and on each of the five seeds, with 4.6--304$\times$ fewer training samples (\cref{fig:teaser}a).
The baselines spend their larger budgets on making DMD stable, which does not add the signals that its gradient lacks.
\method supplies them: it has the highest dynamic degree of all models and leads most in spatial relationship, 5.4 points above the best baseline, so its extra motion does not cost layout.
A small budget suffices for appearance, as the static gradient component is consistent (\cref{sec:analysis}), and \method has higher aesthetic and imaging quality than every baseline.
The supplement lists all 16 dimensions.
\method also beats its 50-step teacher, 85.57 against 82.81 on the same seed, and takes 10.6\,s per video on one A100, 16$\times$ faster.

\paragraph{Qualitative comparison.}
\Cref{fig:qualitative} shows the failures behind the numbers on the same prompts: the baselines fuse, lose, or smear subjects or cut to a new shot, while \method completes each motion with every subject intact.
Here SC holds each subject in place, and CE spreads its motion evenly instead of letting it jump.

\subsection{Why plain DMD lags in motion}
\label{sec:analysis}

We measure the gradient of plain DMD at five noise levels, with four noise draws for each of 32 training captions, and split it as in \cref{eq:decomp} (\cref*{sec:supp_grad}).
Its static component has single-draw signal-to-noise ratios of 0.4 to 1.4 and its layout part up to 9.5, whereas the motion component stays at 0.2 to 0.3 at every noise level (\cref{fig:obs}c, top).
A step size that converges appearance within a few hundred iterations therefore moves motion only slowly, and the push toward even motion stays below its noise, so the stutter of \cref{fig:obs}e persists (\cref{fig:obs}c, bottom).

The two corrections supply these missing signals in their own components: 93\% of the energy of $\dce$ lies in the motion component, against 53 to 57\% for the DMD gradient, and $\dsc$ lies in the layout component by construction.
CE is only about 2\% of the DMD gradient in its window, but while the push of the DMD gradient changes sign between draws, CE points the same way in every draw, so our update pushes toward even motion by 1.5 to 3.5 times its noise (\cref{fig:obs}c, bottom).
SC, which does not depend on the noise draw, raises the signal-to-noise ratio of the layout up to 26 times, yet it is a median of only 15\% of the DMD gradient and skips last inputs at $s\geq0.9$, where no layout has been committed yet (\cref*{sec:supp_mech}).
These two mechanisms account for the results of \cref{sec:main}: the student moves more and more evenly, and its objects stay in place.

\subsection{Fewer steps and another backbone}
\label{sec:transfer}

\begin{table}[t]
\centering
\caption{\textbf{Fewer steps and another backbone.} VBench, 5 videos per prompt (seeds 42--46; teacher: 1), mean $\pm$ standard deviation over the seeds. \emph{Top}: Wan2.1-1.3B in three steps, FastWan's setting (ours without retraining). \emph{Bottom}: Cosmos-Predict2.5-2B, 832$\times$480, 93 frames.}
\label{tab:transfer}
\scalebox{0.8}{%
\begin{tabular}{lrccc}
\toprule
Method & NFE & Total & Quality & Semantic \\
\midrule
\multicolumn{5}{l}{\emph{Wan2.1-T2V-1.3B, 3 steps}} \\
FastWan~\cite{vsa} & 3 & 83.91{\scriptsize$\pm$0.30} & 85.49{\scriptsize$\pm$0.40} & 77.55{\scriptsize$\pm$0.25} \\
\method & 3 & \textbf{85.23}{\scriptsize$\pm$0.22} & \textbf{86.20}{\scriptsize$\pm$0.18} & \textbf{81.35}{\scriptsize$\pm$0.45} \\
\midrule
\multicolumn{5}{l}{\emph{Cosmos-Predict2.5-2B}} \\
Teacher (35 steps) & 70 & 83.20 & 84.54 & 77.82 \\
NVIDIA 4-step~\cite{cosmos25} & 4 & 83.17{\scriptsize$\pm$0.44} & 83.40{\scriptsize$\pm$0.41} & 82.27{\scriptsize$\pm$0.78} \\
\method & 4 & \textbf{83.83}{\scriptsize$\pm$0.29} & \textbf{84.87}{\scriptsize$\pm$0.45} & 79.68{\scriptsize$\pm$0.51} \\
\bottomrule
\end{tabular}}
\vspace{-1em}
\end{table}

\paragraph{Three steps.}
FastWan is distilled for three steps.
Sampled in three steps without retraining, which works because training already includes three-step rollouts, \method leads FastWan by 1.32 points and on every seed (\cref{tab:transfer}, top).
It even beats every 4-step model of \cref{tab:main}, so \method stays effective with fewer sampling steps.

\paragraph{Cosmos-Predict2.5-2B.}
We apply the recipe unchanged, with all training hyperparameters and 600 iterations, to the post-trained Cosmos-Predict2.5-2B~\cite{cosmos25} at 832$\times$480 and 93 frames.
No real video is used: 9{,}216 captions and 72 teacher samples, whose latents only set the shape and CE's reference.
\method outperforms both NVIDIA's released 4-step model, run with its own sampler, and the 35-step teacher, which takes 70 network evaluations (\cref{tab:transfer}, bottom).
The lead comes from motion and frame quality, with a dynamic degree of 74 against 63 and an imaging quality of 71.5 against 67.9, and since Cosmos uses the Wan2.1 VAE, CE also keeps its chunk stutter 45\% below that of NVIDIA's model.
The recipe thus carries over to a different transformer and text encoder without retuning, as both corrections act on the decoded frames and on the student's own last step, not on the architecture.

\subsection{Ablations}
\label{sec:ablation}

\begin{table}[t]
\centering
\caption{\textbf{Ablations on Wan2.1-T2V-1.3B.} VBench, one video per prompt (seed 42), hence no $\pm$. All variants share the minimal setup of \cref{sec:obs}. Full-band SC: no block means.}
\label{tab:ablation}
\resizebox{\linewidth}{!}{%
\begin{tabular}{lccc}
\toprule
Variant & Total & Quality & Semantic \\
\midrule
\multicolumn{4}{l}{\emph{Objective and components, 600 iterations}} \\
rCM objective (sCM + DMD) & 83.21 & 84.73 & 77.15 \\
Plain DMD & 84.62 & 85.93 & 79.39 \\
\;+ CE & 84.66 & 86.40 & 77.67 \\
\;+ SC & 84.07 & 85.09 & 79.97 \\
\;+ CE + full-band SC & 84.35 & 85.80 & 78.57 \\
\;+ CE + SC (\method) & \textbf{85.57} & \textbf{86.52} & \textbf{81.81} \\
\midrule
\multicolumn{4}{l}{\emph{300 iterations}} \\
Plain DMD & 80.99 & 81.91 & 77.30 \\
\method & \textbf{81.78} & \textbf{82.27} & \textbf{79.82} \\
\bottomrule
\end{tabular}}
\vspace{-1em}
\end{table}

\Cref{tab:ablation} ablates the objective, the two corrections, the block averaging of SC, and the training length, all in the minimal setup of \cref{sec:obs}.
(1)~\emph{The setup alone is not enough.}
With the objective of rCM, the same setup scores 1.41 points below plain DMD, so the plain DMD objective matters as well.
(2)~\emph{The two corrections need each other.}
CE alone frees the motion and raises the quality score, but without a layout signal the layout drifts and the semantic score drops by 1.7 points.
SC alone keeps the layout and raises the semantic score, but it adds no motion, so the quality score does not rise.
Together, CE moves the student and SC holds the layout in a component orthogonal to the motion, so the full model gives the best quality and semantic scores and the least stutter.
Over the five seeds of the main comparison, it gains 1.13 points over plain DMD, which alone already scores above every released model.
(3)~\emph{SC must be coarse.}
With a full-band anchor, the student copies the teacher's blurry one-step details, and the total falls by 1.22 points.
(4)~\emph{Faster convergence.}
At 300 iterations, \method already leads plain DMD in every score, most in semantics, because SC's layout signal does not depend on the noise draw, so the corrections speed up learning.
The supplement shows the drift of CE alone and all stutter statistics.

\section{Conclusion}
\label{sec:conclusion}

We have presented \method, a few-step video distillation method that decouples the correction of motion from that of layout.
We found that the DMD gradient learns per-frame quality within a few hundred iterations but carries little signal for even motion within a latent chunk and none for keeping the layout in the last sampling step.
\method supplies these two signals with Chunk Evenness and Semantic Commitment, two training-only corrections of the DMD target.
At a fraction of the training cost of prior work, it gives the best few-step Wan2.1-T2V-1.3B model on VBench, and it remains effective with fewer sampling steps and on another backbone without retuning.

\section*{LLM Usage Statement}

OpenAI's ChatGPT was used to assist with language editing and polishing of the manuscript, as well as code editing and refinement. The authors take full responsibility for the technical content, experimental results, and correctness of the final submission.

\clearpage
\setcounter{figure}{0}
\setcounter{table}{0}
\setcounter{equation}{0}
\setcounter{algorithm}{0}
\renewcommand{\thefigure}{S\arabic{figure}}
\renewcommand{\thetable}{S\arabic{table}}
\renewcommand{\theequation}{S\arabic{equation}}
\renewcommand{\thealgorithm}{S\arabic{algorithm}}
\crefname{appendix}{Sec.}{Secs.}\Crefname{appendix}{Section}{Sections}
\maketitlesupplementary
\appendix

\section{Implementation details}
\label{sec:supp_impl}

\Cref{tab:supp_hparams} lists every hyperparameter of \method.
\providecommand{\suppcodesentence}{The supplementary material contains the code of \method.}%
\suppcodesentence{}
Both backbones use the same training values.
In both, the teacher guidance is the default for distilling that teacher: $w=5$ for Wan2.1, as in rCM~\cite{rcm}, and guidance 3 in Cosmos's convention ($w=4$) for Cosmos, as in NVIDIA's DMD2 recipe for Predict2.5~\cite{cosmos25}.
Only the shapes that follow from the video size, the data, and the first sampling time at inference differ.
Distilled instead with guidance 7 ($w=8$), the default of the 35-step Cosmos sampler, \method scores 83.89$\pm$0.24 on VBench (five seeds, as in \cref*{tab:transfer}), above NVIDIA's 4-step model on every seed and above the teacher, so its lead does not depend on this choice.

\begin{table*}[t]
\centering
\caption{\textbf{Hyperparameters of \method.} One column applies to both backbones unless a second value is given (Cosmos-Predict2.5-2B).}
\label{tab:supp_hparams}
\small
\begin{tabular}{ll}
\toprule
\multicolumn{2}{l}{\emph{Optimization}} \\
Optimizer & AdamW, $\beta=(0, 0.999)$, $\epsilon=10^{-8}$, weight decay 0.01 \\
Student / critic learning rate & $2{\times}10^{-5}$ / $4{\times}10^{-6}$ \\
Learning-rate schedule & linear warm-up over the first 100 updates of each network (200 iterations), then constant \\
Update schedule & critic and student alternate (1:1) \\
Iterations & 600 (300 student and 300 critic updates) \\
Batch & 14 = 7 GPUs $\times$ 1 video $\times$ 2 accumulation steps \\
Gradient clipping & none, as in rCM~\cite{rcm} \\
EMA of the student & power-function EMA, rate 0.1; all results use EMA weights \\
Precision / sharding & bf16 weights and AdamW states, fp32 EMA and gradient reduction; FSDP over 7 GPUs \\
Initialization & student and critic from the teacher; no initialization stage \\
\midrule
\multicolumn{2}{l}{\emph{DMD}} \\
Teacher guidance $w$ & 5 with Wan's negative prompt / 4 (guidance 3 in Cosmos's convention, \\
& as in NVIDIA's DMD2) with its default negative prompt \\
Training rollout & $k=1,2,3,4$ steps in turn from pure noise ($t=\pi/2$); \\
& each intermediate time drawn from $p(t)$ and capped at the previous one; \\
& gradient through the last step only \\
Noise times $p(t)$ & uniform in rectified-flow time, shifted with shift 5 \\
Normalizer $\omega$ & mean absolute difference $\lVert\xG-\xreal\rVert_1/N$ (\cref*{eq:dmd}) \\
Inference times (TrigFlow) & $(\arctan 1600, 1.5, 1.4, 1.0)$; 3 steps: $(\arctan 1600, 1.5, 1.4)$ \\
& Cosmos: $(\arctan 80, 1.5, 1.4, 1.0)$ \\
\midrule
\multicolumn{2}{l}{\emph{Chunk Evenness}} \\
Window & $L=6$ latent frames $\times$ 16$\times$16 latent pixels (21 frames of 128$\times$128) \\
Data reference $\Edata$ & EMA (decay 0.98 per accumulation step) of $E$ on training latents, averaged over GPUs \\
Margin $m$ & 0.2 \\
Normalized gradient steps & $\le 4$, step size $\eta=0.03$; keep motion and detail $\ge 95\%$; clip $\pm 0.5$ \\
\midrule
\multicolumn{2}{l}{\emph{Semantic Commitment}} \\
Weight $\lambda$ / threshold $\tau$ & 1 / 0.9 (on the rectified-flow time $s$ of the last input) \\
Block grid & 12$\times$13 blocks of 5$\times$8 latent pixels \\
Time average & over all latent frames \\
Cap & $\lVert\dsc\rVert_2\le 0.5\,\lVert g-\dce\rVert_2$ per sample \\
\midrule
\multicolumn{2}{l}{\emph{Data and video size}} \\
Captions & rCM synthetic Wan2.1-14B set~\cite{rcm} (250K clips) / 9{,}216 of its captions \\
Latents for $\Edata$ & the same clips / 72 samples of the 35-step teacher \\
Video & 832$\times$480, 81 frames (21 latents) / 832$\times$480, 93 frames (24 latents) \\
\midrule
\multicolumn{2}{l}{\emph{Cost}} \\
Time per iteration & 43.6\,s / 64.8\,s on 7 A100 GPUs \\
Total & 51 / 76 A100-hours \\
\bottomrule
\end{tabular}
\end{table*}

\paragraph{Rollouts and the critic.}
Each student update rolls the student out from pure noise for $k$ steps, cycling through $k=1,\dots,4$.
Each intermediate time is drawn from the noise distribution $p(t)$ and capped at the previous one, as in rCM~\cite{rcm}, and gradients flow only through the last step.
The critic is a copy of the teacher trained with the denoising loss on its own student rollouts, generated without gradients, one update per student update.
Over 600 iterations, the student and the critic each see 4.2k rollouts.
\Cref{alg:update} lists one student update of \method.

\begin{algorithm}[t]
\caption{One student update of \method.}
\label{alg:update}
\begin{algorithmic}[1]
\State Roll out $k\in\{1,\dots,4\}$ student steps from noise; keep the last input $(x_{\mathrm{in}},t_{\mathrm{in}})$ and the output $\xG$
\State $t\sim p(t)$,\; $x_t\gets\cos t\,\xG+\sin t\,\epsilon$
\State $g\gets(\xfake-\xreal)/\omega$ \Comment{DMD}
\State $\dce\gets$ normalized gradient steps on $E$ in a random window \Comment{CE}
\State $\dsc\gets\lambda\,\bar P(\xG-A)/\omega$ if $k\geq2$ and $s(t_{\mathrm{in}})<\tau$, capped; else $0$ \Comment{SC}
\State $\xt\gets\sg(\xG-g+\dce-\dsc)$
\State Update $\theta$ with $\nabla_\theta\,\tfrac12\lVert\xG-\xt\rVert_2^2$
\end{algorithmic}
\end{algorithm}

\paragraph{Chunk Evenness.}
Per sample, CE decodes the 6-latent window of the student output once to compute $E$ and its gradient, and at most four more times, once per trial step.
Each of these decodes includes a backward pass through the decoder.
The window starts at a random latent frame other than the first and at a random spatial position; frames are converted to grayscale by averaging the color channels, and the first decoded frame, which the causal decoder renders as a still image, is dropped.
Windows whose evenness is within $1+m=1.2$ times the data level are left untouched.
The accepted displacement is clipped to $\pm0.5$ and set to zero on the window's first latent frame, whose gradient is that of a still image.
$\Edata$ is updated at each of the two accumulation steps of a student update from one random window of the training latents on each GPU, decoded without gradients, and averaged across GPUs, so all workers use the same reference.

\paragraph{Semantic Commitment.}
The anchor of \cref*{eq:anchor} is one guided teacher prediction (two network evaluations) from the input of the student's last step, computed without gradients.
For $k=1$ the input is pure noise and SC is off.
The block grid divides the 60$\times$104 latent frame exactly.

\paragraph{Cosmos-Predict2.5-2B.}
We wrap the diffusers Cosmos transformer as the student, teacher and critic; the post-trained 2B checkpoint is the teacher.
Its text embeddings come from Cosmos's own encoder and are computed once for the 9{,}216 captions.
Cosmos uses the Wan2.1 VAE, so CE runs unchanged.
The 72 teacher samples (35 steps) only set the latent shape and CE's reference $\Edata$; DMD itself needs no video.

\section{Evaluation details}
\label{sec:supp_eval}

\paragraph{VBench protocol.}
All models generate the 946 GPT-4o-augmented VBench prompts; video $i$ of a prompt uses seed $42+i$ ($i=0,\dots,4$).
Two prompts occur twice in VBench's list, for different dimensions; since VBench names a video by its prompt, each seed has 944 distinct videos for every model, and both entries of a repeated prompt score the same video.
Videos are written with libx264 (CRF 18) at 16 frames per second and scored with the official VBench code and weights in its standard mode, on all 16 dimensions.
The quality score averages the seven quality dimensions and the semantic score the nine semantic dimensions with VBench's normalization and weights; the total is $(4\,\text{quality}+\text{semantic})/5$.
Ablations use seed 42 only.
The $\pm$ in \cref*{tab:main,tab:transfer} is the standard deviation of each score over the five seeds, each scored as its own 946-prompt set.

\begin{table}[t]
\centering
\caption{\textbf{Teachers and released students.} The teachers serve as references, and the released students are the baselines of \cref*{tab:main,tab:transfer}. All run with their official samplers through our pipeline (832$\times$480, 81 frames; Cosmos: 93 frames).}
\label{tab:supp_baselines}
\resizebox{\linewidth}{!}{%
\begin{tabular}{lll}
\toprule
Method & Checkpoint & Sampling \\
\midrule
\multicolumn{3}{l}{\emph{Teachers (references)}} \\
Wan2.1 teacher & Wan2.1-T2V-1.3B & 50-step UniPC, $w=5$ (100 NFE) \\
Cosmos teacher & Predict2.5-2B post-trained & official 35 steps, guidance 7 (70 NFE) \\
\midrule
\multicolumn{3}{l}{\emph{Released students (baselines)}} \\
FastWan~\cite{vsa} & FastWan2.1-T2V-1.3B & official 3-step sampler, dense attention \\
CoDMD~\cite{codmd} & CoDMD Wan2.1-T2V-1.3B & official 4-step sampler \\
rCM~\cite{rcm} & rCM Wan2.1-T2V-1.3B 480p & 4 steps, TrigFlow $(\arctan 1600, 1.5, 1.4, 1.0)$ \\
Self-Forcing~\cite{selfforcing} & self-forcing DMD (EMA) & 3 latents per block, 4 steps, KV cache \\
NVIDIA 4-step~\cite{cosmos25} & Predict2.5-2B distilled & NVIDIA's 4-step sampler \\
\bottomrule
\end{tabular}}
\end{table}

\paragraph{Baselines.}
\Cref{tab:supp_baselines} lists the checkpoints and samplers.
FastWan runs with dense attention in place of its sparse attention kernel; all other models run as released.
Our students use the rCM sampler with the times in \cref{tab:supp_hparams}, on Cosmos as well.
Both teachers run with their official settings: Wan2.1 with 50 steps and $w=5$, Cosmos with 35 steps and guidance 7, \ie $w=8$ for the guidance scale $w$ of \cref*{sec:prelim}.

\paragraph{Timing.}
Inference times are measured on one A100-80GB in bf16 for one 832$\times$480, 81-frame video, averaged over repeated runs, and split into text encoding, transformer and VAE decoding (\cref{tab:supp_cost}).
Training times are averaged over iterations 2--21 on the seven training GPUs.

\paragraph{Teaser figure.}
In \cref*{fig:teaser}, filled markers are released checkpoints that we generate and score with the protocol above.
Hollow markers are the scores reported in Table~1 of PDMD~\cite{pdmd}, which evaluates all of them under one protocol; we shift each score by the difference between our rCM result and the rCM result there, so that the two protocols agree on rCM.
Training samples are optimizer iterations of student and critic times the global batch, taken from the respective papers (for AnyFlow, its second-stage batch is assumed equal to its first); ADV publishes no budget and appears only in panel (b), and the sample count and cost of Self-Forcing (autoregressive) exclude its initialization stage.
In panel (c) and in the A100-hours of \cref*{tab:main}, the cost of \method is measured, that of CoDMD is its reported 32 A100s for 1.5 days, and the others are their published sample budgets times our measured A100 cost per sample for the same update schedule (\cref{tab:supp_cost}).

\section{Full results}
\label{sec:supp_full}

\paragraph{All VBench dimensions.}
\Cref{tab:supp_dims} expands \cref*{tab:main} to the 16 VBench dimensions.
Compared with plain DMD, \method gains 14.2 points of dynamic degree, 5.5 of spatial relationship, 3.3 of multiple objects and 2.8 of object class, and also improves subject consistency by 1.5 points even though its videos move much more.

\begin{table*}[t]
\centering
\caption{\textbf{All 16 VBench dimensions} on Wan2.1-T2V-1.3B (946 prompts, 5 videos each; teacher: 1 video, seed 42).}
\label{tab:supp_dims}
\small
\begin{tabular}{lccccccc}
\toprule
Dimension & Teacher & FastWan & CoDMD & rCM & Self-Forcing & Plain DMD & \method \\
\midrule
Subject consistency & 92.92 & 95.95 & 96.05 & 93.87 & 94.70 & 95.21 & 96.68 \\
Background consistency & 94.48 & 94.41 & 94.99 & 94.92 & 95.25 & 95.47 & 94.53 \\
Temporal flickering & 99.14 & 99.11 & 98.03 & 98.54 & 98.58 & 98.71 & 98.37 \\
Motion smoothness & 97.72 & 98.42 & 98.48 & 97.52 & 98.11 & 98.92 & 98.49 \\
Dynamic degree & 69.44 & 78.89 & 80.28 & 84.72 & 71.39 & 76.11 & 90.28 \\
Aesthetic quality & 65.71 & 66.41 & 66.64 & 66.81 & 67.55 & 68.27 & 67.96 \\
Imaging quality & 67.77 & 69.12 & 68.45 & 67.30 & 69.72 & 69.91 & 70.37 \\
\midrule
Object class & 91.93 & 92.93 & 93.62 & 94.59 & 94.81 & 94.65 & 97.47 \\
Multiple objects & 70.50 & 75.24 & 80.78 & 84.54 & 80.85 & 81.97 & 85.27 \\
Human action & 96.00 & 96.80 & 95.60 & 97.80 & 97.80 & 94.20 & 95.40 \\
Color & 86.04 & 84.45 & 85.49 & 84.92 & 87.13 & 83.01 & 85.89 \\
Spatial relationship & 79.62 & 73.83 & 73.19 & 81.26 & 82.91 & 82.82 & 88.32 \\
Scene & 56.76 & 49.80 & 52.88 & 52.53 & 54.11 & 53.11 & 54.85 \\
Appearance style & 22.76 & 20.92 & 21.40 & 21.85 & 22.05 & 20.25 & 21.16 \\
Temporal style & 26.08 & 24.99 & 25.04 & 24.97 & 25.49 & 24.98 & 24.85 \\
Overall consistency & 26.88 & 26.33 & 26.20 & 26.63 & 26.98 & 25.80 & 26.04 \\
\midrule
Quality score & 83.56 & 85.49 & 85.26 & 84.74 & 84.74 & 85.87 & \textbf{86.69} \\
Semantic score & 79.81 & 77.55 & 78.74 & 80.58 & 81.18 & 79.07 & \textbf{81.43} \\
Total & 82.81 & 83.91 & 83.95 & 83.91 & 84.03 & 84.51 & \textbf{85.64} \\
\bottomrule
\end{tabular}
\end{table*}

\paragraph{Per-seed totals.}
\Cref{tab:supp_seeds} lists the total score of every seed.
\method is the best model on each of the five seeds, by 1.21 to 2.17 points over the released models and by 0.63 to 1.46 points over plain DMD.

\begin{table}[t]
\centering
\caption{\textbf{VBench total per seed} on Wan2.1-T2V-1.3B (946 prompts per seed).}
\label{tab:supp_seeds}
\resizebox{\linewidth}{!}{%
\begin{tabular}{lccccc|c}
\toprule
Method & 42 & 43 & 44 & 45 & 46 & Mean \\
\midrule
FastWan~\cite{vsa} & 83.66 & 83.56 & 83.92 & 84.25 & 84.15 & 83.91 \\
CoDMD~\cite{codmd} & 84.11 & 83.84 & 84.23 & 83.67 & 83.90 & 83.95 \\
rCM~\cite{rcm} & 83.68 & 83.94 & 83.68 & 84.15 & 84.10 & 83.91 \\
Self-Forcing~\cite{selfforcing} & 84.10 & 83.82 & 84.19 & 84.14 & 83.90 & 84.03 \\
Plain DMD & 84.62 & 84.31 & 84.81 & 84.41 & 84.39 & 84.51 \\
\method & \textbf{85.57} & \textbf{85.73} & \textbf{85.44} & \textbf{85.58} & \textbf{85.85} & \textbf{85.64} \\
\bottomrule
\end{tabular}}
\end{table}

\paragraph{Cost.}
\Cref{tab:supp_cost} breaks down inference and training time.
\method generates exactly like rCM and CoDMD, 16$\times$ faster than the teacher; in three steps it takes 9.0\,s, the time of FastWan with dense attention.
In training, CE and SC add 9\% to an iteration of plain DMD.
Because the critic is updated once per student update instead of four times, a student update of \method costs 87\,s, against 180\,s for a DMD2-style schedule and 225\,s for the rCM recipe at the same batch.

\begin{table}[t]
\centering
\caption{\textbf{Cost.} \emph{Top}: one 832$\times$480, 81-frame video on one A100 (seconds). \emph{Bottom}: training on 7 A100 GPUs, batch 14.}
\label{tab:supp_cost}
\resizebox{\linewidth}{!}{%
\begin{tabular}{lccccc}
\toprule
Inference & NFE & Transformer & VAE & Total & Memory (GiB) \\
\midrule
Wan2.1 teacher & 100 & 165.9 & 4.0 & 170.0 & 21.0 \\
Self-Forcing & 35 & 7.3 & 3.2 & 10.5 & 23.7 \\
FastWan (dense) & 3 & 5.0 & 3.9 & 9.0 & 21.0 \\
rCM / CoDMD / \method & 4 & 6.6 & 4.0 & 10.6 & 21.0 \\
\midrule
Training & \multicolumn{2}{c}{Critic : student} & s / iteration & \multicolumn{2}{c}{s / student update} \\
\midrule
DMD2-style schedule & \multicolumn{2}{c}{4 : 1} & 36.1 & \multicolumn{2}{c}{180} \\
rCM recipe (sCM + DMD) & \multicolumn{2}{c}{4 : 1} & 44.9 & \multicolumn{2}{c}{225} \\
Plain DMD (our recipe) & \multicolumn{2}{c}{1 : 1} & 40.1 & \multicolumn{2}{c}{80} \\
\method & \multicolumn{2}{c}{1 : 1} & 43.6 & \multicolumn{2}{c}{87} \\
\bottomrule
\end{tabular}}
\end{table}

\paragraph{Guidance scale.}
FastWan and CoDMD distill with guidance 3.5.
Plain DMD with $w=3.5$ scores 84.58 (quality 85.97, semantic 79.00) against 84.62 (85.93, 79.39) with $w=5$ (seed 42), and still leads FastWan by 0.92 points; the guidance scale does not explain the gap.

\paragraph{rCM's objective in the minimal setup.}
To separate the setup from the objective, we train the objective of rCM (sCM + DMD, sCM weight 100 as in rCM) in the minimal setup of \cref*{sec:obs}: student initialized from the teacher, one critic update per student update, our learning rates, and 600 iterations, without rCM's sCM-only warmup of 1000 iterations, which would exceed the budget.
Training is stable, but the student scores 83.21, 1.41 points below plain DMD and 2.36 below \method (\cref{tab:supp_rcmobj}; also in \cref*{tab:ablation}): aesthetic quality drops from 68.2 to 62.3 and the semantic score from 79.4 to 77.2, while the dynamic degree rises from 76 to 83.
The learning rate alone thus does not make a strong student, and the gain of \method comes on top of plain DMD, the stronger of the two objectives in this setup.

\begin{table}[t]
\centering
\caption{\textbf{Objective versus setup.} All three are trained in the minimal setup of \cref*{sec:obs} (teacher initialization, one critic update per student update, our learning rates, 600 iterations); VBench, seed 42.}
\label{tab:supp_rcmobj}
\resizebox{\linewidth}{!}{%
\begin{tabular}{lccc}
\toprule
Objective & Total & Quality & Semantic \\
\midrule
rCM (sCM + DMD) & 83.21 & 84.73 & 77.15 \\
Plain DMD & 84.62 & 85.93 & 79.39 \\
\method (DMD + CE + SC) & \textbf{85.57} & \textbf{86.52} & \textbf{81.81} \\
\bottomrule
\end{tabular}}
\end{table}

\section{Stutter and motion metrics}
\label{sec:supp_metrics}

VBench measures how much a video moves (dynamic degree) and how smooth it is overall (motion smoothness, temporal flickering), but not how the motion is distributed over the four frames that one latent frame decodes into.
We therefore use a few statistics of the decoded frames.
All of them start from the same frame-difference sequence, and none of them is used for training.

\paragraph{Frame-difference sequence.}
We convert a video to grayscale at half resolution and split every frame into 64$\times$64 blocks (128 pixels at full resolution).
For each block, $r_j$ is the RMS difference between frames $j$ and $j{+}1$.
Stutter is visible only where things move, so we keep the two blocks with the largest mean difference and average each statistic over them.

\paragraph{Stutter amplitude $A_4$.}
Evenly spread motion gives a flat sequence $r_j$; motion packed into some frames of each four-frame chunk gives a sawtooth with period four.
We measure its strength by the period-4 Fourier amplitude relative to the mean motion,
\begin{equation}
A_4=\Big|\frac{1}{J}\sum_{j=1}^{J} r_j\,e^{-i\pi j/2}\Big|\Big/\bar r .
\label{eq:a4}
\end{equation}
$A_4$ is zero for evenly spread motion and grows with the sawtooth; dividing by the mean motion $\bar r$ makes it independent of how much the video moves.
We report $100\,A_4$, as the median over videos.
On Wan, $A_4$ is computed on 40 prompts with large motion (seed 42), rendered from the latents without video compression so that the encoder cannot smooth the sawtooth; on Cosmos, on all 944 videos of seed 42 (CRF 18), the same set for both models.

\paragraph{Local evenness and severe stutter.}
To find the worst cases, we evaluate the phase evenness $E$ of \cref*{eq:evenness} on the same two blocks, with the phase means of $r_j$ over $j \bmod 4$.
A video stutters severely when $E>0.1$.

\paragraph{Hand boundary ratio.}
For prompts about hands, the block is the 128-pixel box with the most skin-colored motion.
The boundary ratio divides the mean change of this box at transitions that cross a latent chunk by its mean change at the other transitions.
It is 1 for even motion and grows when the hand moves only at chunk boundaries.

\paragraph{Motion-compensated residual.}
A hand can also change its shape instead of stuttering (\cref{fig:hands_dishes}).
In the hand box, we match every 16$\times$16 block of a frame to the next frame within $\pm6$ pixels and divide the residual after the best match by the raw frame difference.
Rigid or smoothly moving fingers leave a small residual; fingers that change shape leave a large one.

\subsection*{Results}

\paragraph{Ablations.}
\Cref{tab:supp_a4} adds $A_4$ to the ablation of \cref*{tab:ablation}.
Each correction lowers the stutter of plain DMD, and together they bring it below the teacher's level (1.45 against 1.66).
At 300 iterations, when plain DMD stutters at three times the teacher's level (\cref*{fig:obs}e), \method is already at 1.95.

\begin{table}[t]
\centering
\caption{\textbf{Stutter amplitude in the ablations} ($100\,A_4$, median over 40 large-motion prompts, lossless renders, seed 42).}
\label{tab:supp_a4}
\setlength{\tabcolsep}{6pt}
\begin{tabular}{lcc}
\toprule
Variant & 300 it. & 600 it. \\
\midrule
Plain DMD & 5.13 & 2.13 \\
\;+ CE & -- & 1.84 \\
\;+ SC & -- & 1.91 \\
\;+ CE + full-band SC & -- & 1.97 \\
\;+ CE + SC (\method) & \textbf{1.95} & \textbf{1.45} \\
\midrule
Teacher (50 steps) & \multicolumn{2}{c}{1.66} \\
\bottomrule
\end{tabular}
\end{table}

\paragraph{Severe stutter is removed, not just averaged down.}
CE corrects a window only when its unevenness exceeds that of the training clips, so it should remove the worst cases rather than shift the median.
On 395 VBench prompts (seed 42), \method stutters severely in 0.3\% of the videos, against 3.0--9.1\% for the released models and 6.8\% for plain DMD, and has the lowest 90th percentile of $E$ (\cref{tab:supp_severe}).

\begin{table}[t]
\centering
\caption{\textbf{Severe chunk stutter} on 395 VBench prompts (seed 42): share of videos with local evenness $E>0.1$, and the 90th percentile of $E$.}
\label{tab:supp_severe}
\setlength{\tabcolsep}{4pt}
\begin{tabular}{lcc}
\toprule
Method & $E>0.1$ (\%)\,$\downarrow$ & 90th pct.\,$\downarrow$ \\
\midrule
FastWan~\cite{vsa} & 9.1 & 0.093 \\
rCM~\cite{rcm} & 6.1 & 0.058 \\
Self-Forcing~\cite{selfforcing} & 5.8 & 0.074 \\
CoDMD~\cite{codmd} & 3.0 & 0.041 \\
Plain DMD & 6.8 & 0.090 \\
\method & \textbf{0.3} & \textbf{0.027} \\
\bottomrule
\end{tabular}
\end{table}

\paragraph{Hands.}
We measure hand stutter on twelve prompts in which a person works with their hands (clapping, counting money, folding clothes, tying a tie, shuffling cards, and others).
At 300 iterations, the chunk-crossing transitions of plain DMD change the hand 2.5$\times$ (tie) and 1.6$\times$ (clapping) as much as the others, against 0.95$\times$ and 0.82$\times$ for \method.
Over all twelve prompts (\cref{tab:supp_hands}), \method moves hands evenly already at 300 iterations, with a 5.6$\times$ lower $A_4$ than plain DMD, and stays closest to even at 600.

\begin{table}[t]
\centering
\caption{\textbf{Hand stutter} on 12 hand prompts ($100\,A_4$ in the hand box and boundary ratio; median over prompts, lossless renders, seed 42).}
\label{tab:supp_hands}
\setlength{\tabcolsep}{4pt}
\begin{tabular}{lcccc}
\toprule
& \multicolumn{2}{c}{300 iterations} & \multicolumn{2}{c}{600 iterations} \\
\cmidrule(lr){2-3}\cmidrule(lr){4-5}
Method & $A_4\downarrow$ & Ratio & $A_4\downarrow$ & Ratio \\
\midrule
Plain DMD & 10.1 & 1.24 & 5.2 & 0.91 \\
\method & \textbf{1.8} & \textbf{1.00} & \textbf{3.0} & \textbf{0.99} \\
\bottomrule
\end{tabular}
\end{table}

\paragraph{Finger stability.}
In \cref{fig:hands_dishes}, 68\% of the glove's frame-to-frame change remains after motion compensation for the conventional DMD run after 4100 iterations.
For \method it is 50\%, the teacher's level (49\%): its fingers move rather than change shape.

\paragraph{Cosmos.}
Over all 944 VBench videos of seed 42, the median $100\,A_4$ of \method is 5.7 against 10.4 for NVIDIA's 4-step model (45\% lower; lower on 76\% of the videos), while its motion $\bar r$ is 19\% larger (26.5 against 22.3).

\section{What the DMD gradient carries}
\label{sec:supp_grad}

For a checkpoint of plain DMD in the minimal setup of \cref*{sec:obs} and 32 captions of the training set, the student generates $\xG$ with its 4-step sampler.
At each of five rectified-flow noise levels $s$, four noise draws give DMD gradients $g=(\xfake-\xreal)/\omega$ as in training (\cref*{eq:dmd}), with the critic of the same checkpoint and the guided teacher.
We split $g$ into the components of \cref*{eq:decomp}: the static component $Mg$ (the mean over latent frames), its layout $\bar P g$ (block means on the $12{\times}13$ grid of SC), and the motion component $(I-M)g$, and report the single-draw signal-to-noise ratio $\lVert\mathbb{E}g\rVert^2/\mathbb{E}\lVert g-\mathbb{E}g\rVert^2$ of each, estimated without bias from the draws.
On four CE windows per sample (\cref*{sec:ce}), we compute the cosine between $-g$, averaged over the four draws, and $-\nabla_z E$, the direction that lowers the phase evenness $E$ of \cref*{eq:evenness}; for a random direction, this cosine has a standard deviation of $1/\sqrt{d}\approx0.0064$ over the $d=24576$ latent values of a window.

\Cref{tab:supp_grad} shows that the static and layout components carry consistent signal, most of it at low noise, while the motion component stays between 0.2 and 0.3 at every noise level and at both checkpoints.
At 200 iterations, when the stutter peaks (\cref*{fig:obs}e), the cosine with the evenness direction is negative at low noise and positive only at $s=0.8$; weighted by the noise levels DMD trains on, it is $(-0.11\pm0.12)\times10^{-2}$, that is, zero.
At 600 iterations it is positive at every level, $(+0.53\pm0.13)\times10^{-2}$ after weighting, below the standard deviation of a random direction.
Single draws reach $|\cos|$ of 1.4 to 3.6 times that of a random direction at 200 iterations and about twice at 600, with a sign that changes from draw to draw.

\paragraph{Where the corrections act.}
SC's correction $\bar P(\xG-A)$ lies in the layout component by construction.
For CE, we run its correction exactly as in training on 16 samples of plain DMD at 200 iterations, when the stutter peaks, with four windows each, and measure the share of the energy of $\dce$ in the motion component $(I-M)$ of the full video.
Over the 45 windows that CE corrects, the median share is 93\% (90th percentile of the static share: 12\%), against 53 to 57\% for the DMD gradient at 100 to 600 iterations; a random direction confined to the same window would have 95\%.
CE thus moves the student almost only along the motion component, and SC only along the layout component.

\begin{table}[t]
\centering
\caption{\textbf{What the DMD gradient carries.} Plain DMD in the minimal setup; 32 captions, 4 noise draws per level. SNR: single-draw signal-to-noise ratio of each part of $g$. $\cos$: gradient averaged over the draws against the direction that lowers the phase evenness $E$, $\times100$, mean $\pm$ standard error over captions; a random direction has a standard deviation of 0.64. Share: fraction of training draws near each noise level.}
\label{tab:supp_grad}
\resizebox{\linewidth}{!}{%
\begin{tabular}{lcccccc}
\toprule
& & & \multicolumn{3}{c}{SNR} & \\
\cmidrule(lr){4-6}
Iteration & $s$ & Share & Static & Layout & Motion & $100\cos$ \\
\midrule
\multirow{5}{*}{600} & 0.2 & 0.08 & 1.40 & 9.46 & 0.28 & $+0.50\pm0.19$ \\
 & 0.4 & 0.09 & 1.07 & 5.93 & 0.26 & $+0.68\pm0.22$ \\
 & 0.6 & 0.15 & 0.85 & 2.95 & 0.24 & $+0.64\pm0.20$ \\
 & 0.8 & 0.27 & 0.59 & 1.24 & 0.22 & $+0.56\pm0.19$ \\
 & 0.95 & 0.42 & 0.45 & 0.60 & 0.22 & $+0.44\pm0.17$ \\
\midrule
\multirow{5}{*}{200} & 0.2 & 0.08 & 1.58 & 8.72 & 0.21 & $-1.19\pm0.31$ \\
 & 0.4 & 0.09 & 1.34 & 6.32 & 0.22 & $-0.79\pm0.30$ \\
 & 0.6 & 0.15 & 1.02 & 3.16 & 0.22 & $-0.11\pm0.29$ \\
 & 0.8 & 0.27 & 0.71 & 1.26 & 0.21 & $+0.58\pm0.21$ \\
 & 0.95 & 0.42 & 0.44 & 0.55 & 0.20 & $-0.19\pm0.12$ \\
\bottomrule
\end{tabular}}
\end{table}

\paragraph{The update of \method.}
We repeat the measurement with the full update of training, $g-\dce+\dsc$ (\cref*{sec:ce,sec:sc}), on the same samples, noise draws, and windows; $\dce$ and $\dsc$ are computed exactly as in training, and the DMD numbers reproduce \cref{tab:supp_grad}.
The push of an update is its projection onto $\nabla E/\lVert\nabla E\rVert$ in a CE window, so that a positive push lowers $E$; we divide its mean over the noise draws by its standard deviation, average over the four windows, and report mean and standard error over the captions.
\Cref{tab:supp_grad_ours} shows that SC raises the signal-to-noise ratio of the static component up to 6 times and that of the layout up to 26 times, since its correction does not depend on the noise draw.
CE is small, about 2\% of the DMD gradient in its window, yet it raises the push toward even motion from below the noise to 1.5 to 3.5 times the noise, because it points the same way in every draw; weighted by the noise levels of training, the push is 2.16 against 0.40 for plain DMD.
On our own checkpoint at 600 iterations, where the stutter is already at the teacher's level and CE corrects 29\% of the windows, the push of our update is smaller (0.74 after weighting) but still positive, while the DMD gradient alone points the other way ($-0.30$).

\begin{table}[t]
\centering
\caption{\textbf{Our update against the DMD gradient.} Plain DMD at 600 iterations (push also at 200), 32 captions; same samples, noise draws, and windows for DMD and for our update $g-\dce+\dsc$. Static, layout: single-draw signal-to-noise ratio. Push: projection onto the even-motion direction, mean over the draws divided by its standard deviation.}
\label{tab:supp_grad_ours}
\resizebox{\linewidth}{!}{%
\begin{tabular}{lcccccc}
\toprule
& \multicolumn{2}{c}{Static} & \multicolumn{2}{c}{Layout} & \multicolumn{2}{c}{Push} \\
\cmidrule(lr){2-3}\cmidrule(lr){4-5}\cmidrule(lr){6-7}
$s$ & DMD & Ours & DMD & Ours & DMD (200 / 600) & Ours \\
\midrule
0.2 & 1.40 & 8.83 & 9.46 & 250 & $-1.40$ / $+0.78$ & $+3.55\pm0.47$ \\
0.4 & 1.07 & 6.43 & 5.93 & 98.2 & $-0.66$ / $+0.54$ & $+3.32\pm0.47$ \\
0.6 & 0.85 & 4.72 & 2.95 & 40.5 & $-0.02$ / $+0.54$ & $+2.70\pm0.34$ \\
0.8 & 0.59 & 2.27 & 1.24 & 10.5 & $+0.32$ / $+0.37$ & $+2.14\pm0.28$ \\
0.95 & 0.45 & 0.57 & 0.60 & 0.79 & $-0.23$ / $+0.27$ & $+1.46\pm0.19$ \\
\bottomrule
\end{tabular}}
\end{table}

\section{The two observations in detail}
\label{sec:supp_obs}

\paragraph{Frames converge before motion.}
\Cref*{fig:obs} follows plain DMD through training.
Per-frame imaging and aesthetic quality reach their final level by 300 iterations, while the dynamic degree is still 18 (\cref*{fig:obs}b).
The stutter amplitude $A_4$ of \cref{sec:supp_metrics} peaks at 4.6 times the teacher's level at 200 iterations and is still three times it at 300 (\cref*{fig:obs}e), and a single clip shows the sawtooth directly (\cref*{fig:obs}d).

\begin{figure*}[t]
\centering
\includegraphics[width=0.92\textwidth]{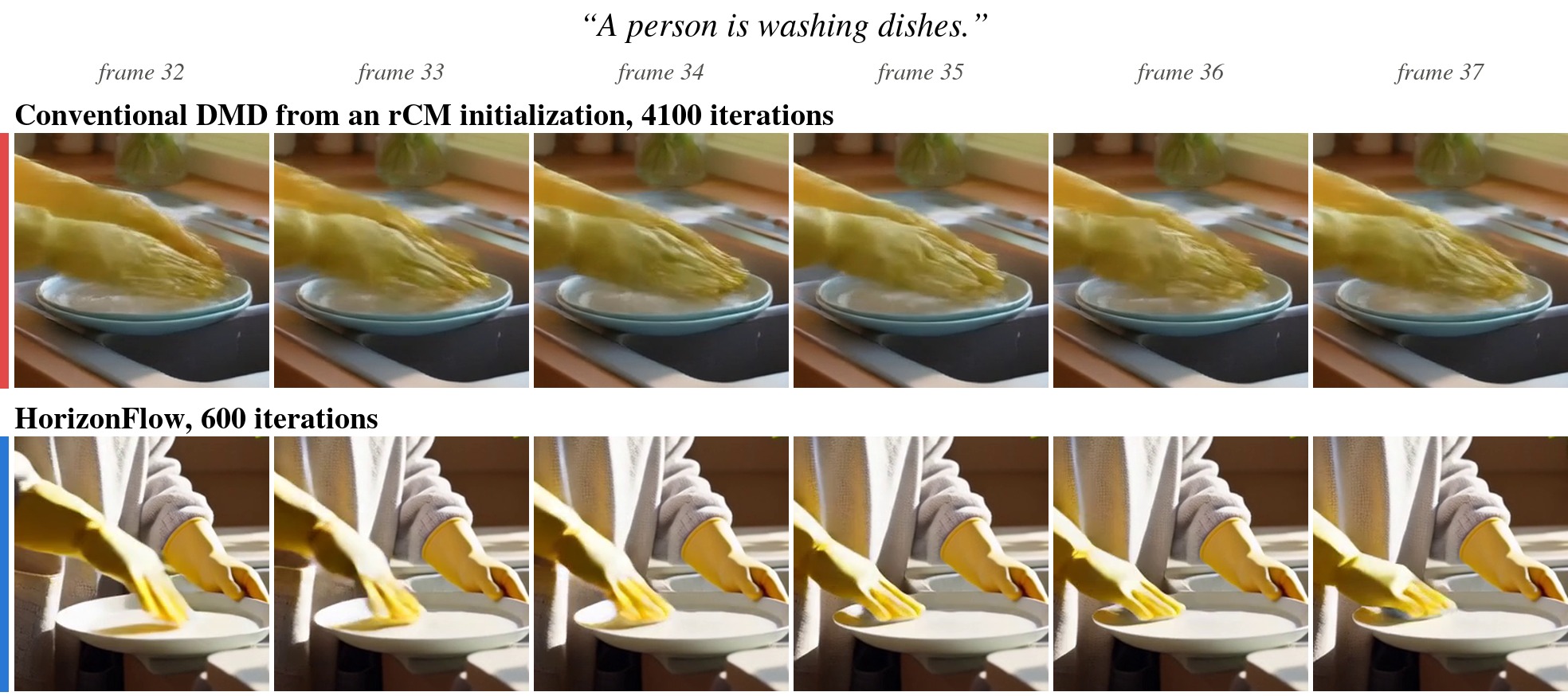}
\caption{\textbf{Fingers under slow training.}
Six consecutive frames of ``A person is washing dishes.''
After 4100 iterations in the conventional regime (initialized from rCM, small learning rate, critic-heavy schedule), the fingers of the glove lose their shape and re-form in every frame.
\method, after 600 iterations, keeps the glove and its fingers intact while the hand wipes the plate.}
\label{fig:hands_dishes}
\end{figure*}

\paragraph{Fine structures.}
\Cref{fig:hands_dishes} shows all six consecutive frames of the example in \cref*{fig:obs}a.

\begin{figure*}[t]
\centering
\includegraphics[width=\textwidth]{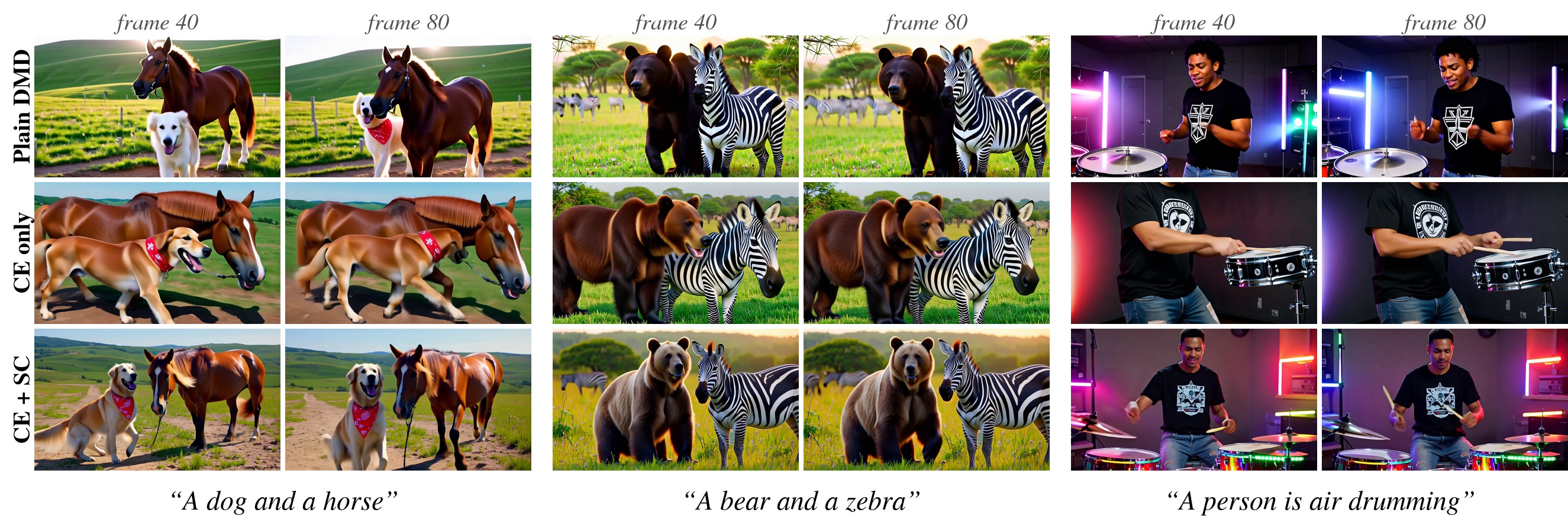}
\caption{\textbf{Unlocked motion rewrites the layout.}
With CE alone, the student moves (dynamic degree 76$\rightarrow$97), but its last step also changes the layout: the dog blends into the horse, the bear's body runs into the zebra's, and the drummer is framed so tightly that the head is cut off.
Semantic Commitment keeps the committed layout while the motion stays.
Frames 40 and 80 after 600 iterations, same seed.}
\label{fig:drift}
\end{figure*}

\paragraph{Unlocked motion rewrites the layout.}
\Cref{fig:drift} compares plain DMD, CE alone, and CE with SC on the same prompts and seed.
CE alone raises the dynamic degree from 76 to 97 and lowers multi-object accuracy from 84 to 76; with SC, the subjects stay separate and in frame while they move.

\section{Mechanism analysis}
\label{sec:supp_mech}

\paragraph{What Semantic Commitment does during training.}
We log SC at every tenth student update of the Wan run that reaches it (68 logged updates).
SC is active in 76\% of the rollouts with $k\geq2$; the remaining ones start their last step at $s\geq0.9$, where the coarse layout of the student and of the teacher's completion differ by 52\% (relative block-mean distance), \ie no layout has been committed yet, and the threshold correctly skips them.
When SC is active, the student's output and the teacher's completion differ in coarse layout by a median of only 3.2\%, and the correction is a median of 15\% of the DMD gradient, never above the cap of 50\%.
SC therefore steers the update without replacing it.

\section{Additional qualitative results}
\label{sec:supp_qual}
\begin{figure*}[p]
\centering
\includegraphics[width=\textwidth,height=0.84\textheight,keepaspectratio]{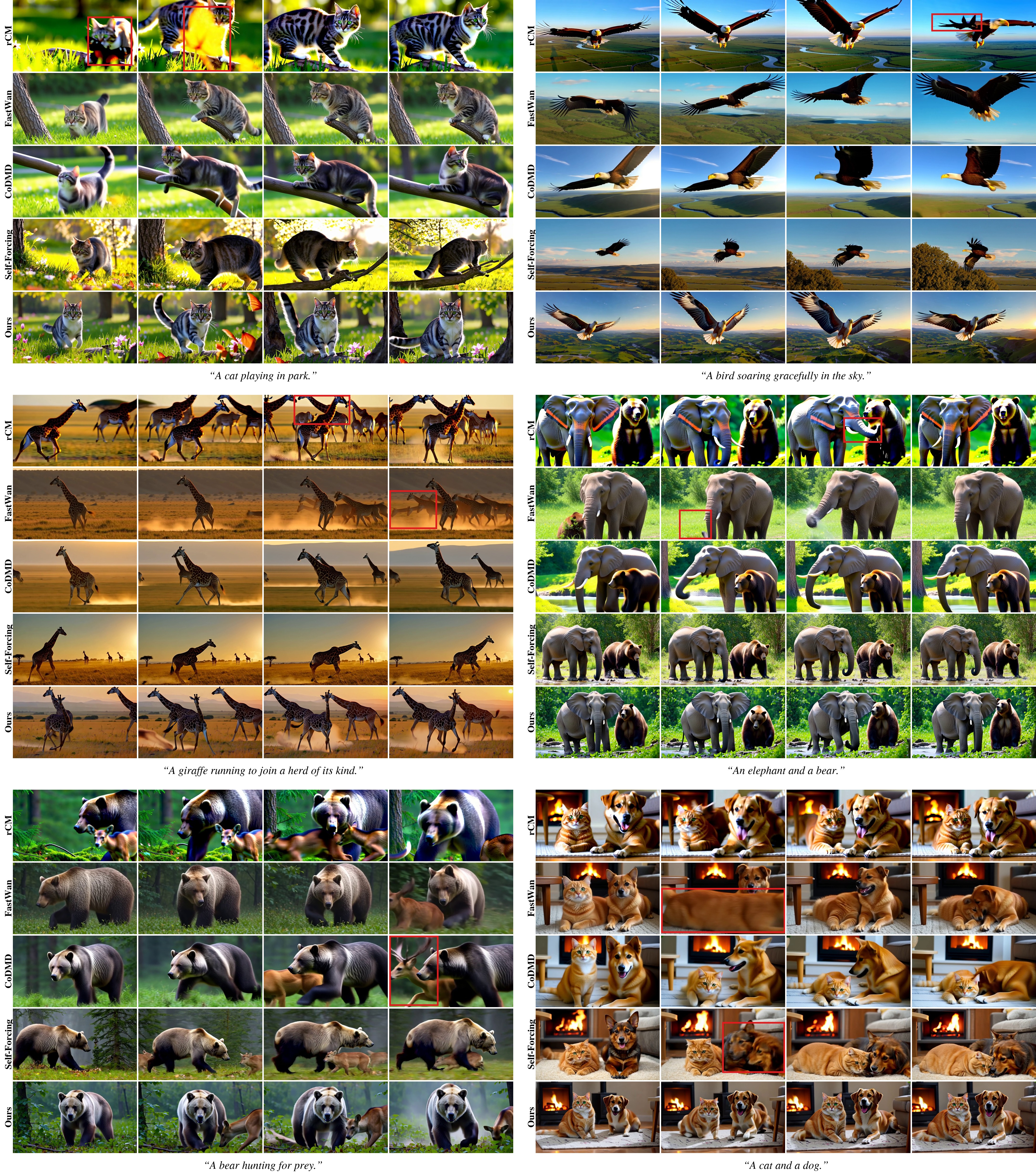}
\caption{\textbf{More comparisons on Wan2.1-T2V-1.3B: animals in motion.}
Every method gets the same VBench prompt and seed; columns are frames 1, 28, 55 and 81 of 81; red boxes mark failures.
rCM fuses the cat with a dark blob and then hides it under a yellow smear, breaks the bird's wing into black shards, tangles the giraffes' necks, and merges the elephant's trunk into the bear.
FastWan renders the giraffe herd as translucent ghosts, loses the bear next to the elephant, and smears the cat across the whole frame.
CoDMD grows a deer out of the bear.
Self-Forcing turns the dog next to the cat into two fused heads.
\method keeps every animal whole and separate while it moves.}
\label{fig:supp_qual_a}
\end{figure*}

\begin{figure*}[p]
\centering
\includegraphics[width=\textwidth,height=0.84\textheight,keepaspectratio]{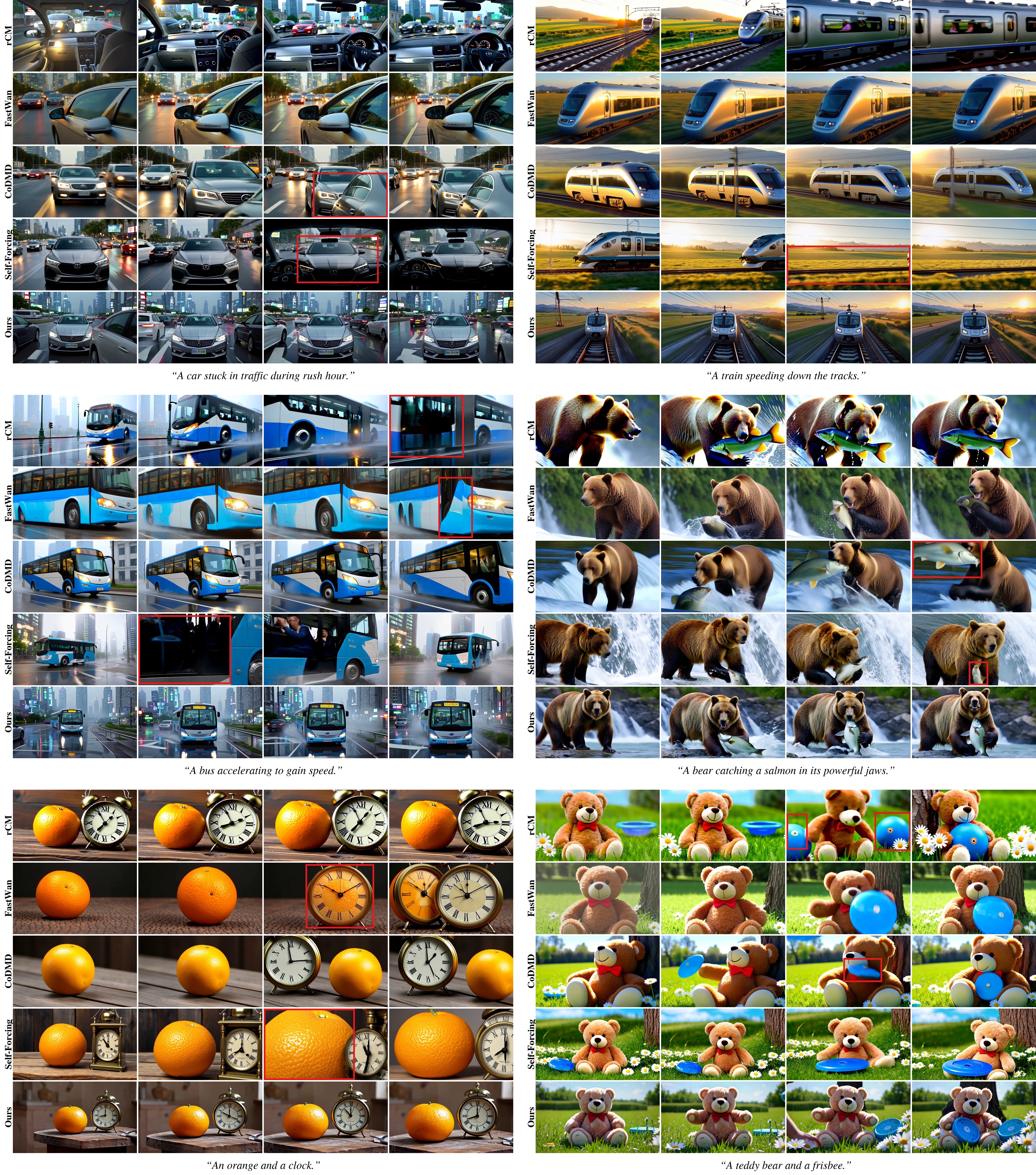}
\caption{\textbf{More comparisons on Wan2.1-T2V-1.3B: traffic, actions and object pairs.}
Same protocol as \cref{fig:supp_qual_a}.
Self-Forcing puts the front of a car inside the cabin, and CoDMD merges two cars.
On the accelerating bus, Self-Forcing cuts to a dark cabin interior, rCM dissolves the side of the bus into dark streaks, and FastWan folds its front panel.
Self-Forcing's train leaves the shot, and the clip ends on an empty field.
CoDMD covers the bear's head with the salmon, and Self-Forcing shrinks the salmon to a sliver.
FastWan turns the orange into a clock, Self-Forcing inflates it until it fills the frame, rCM turns the frisbee into two balls with eyes, and CoDMD fuses the frisbee into the bear's face.
\method keeps the requested subjects in the shot and in one piece, and keeps the two objects of each pair apart.}
\label{fig:supp_qual_b}
\end{figure*}

\Cref{fig:supp_qual_a,fig:supp_qual_b} extend \cref*{fig:qualitative} to twelve more VBench prompts and add CoDMD~\cite{codmd}.
The protocol is the same: every method gets the same prompt and seed, and all videos come from the five-seed sets behind \cref*{tab:main}.
The prompts cover animals in motion, traffic scenes and two-object prompts.
They show the failure modes of \cref*{fig:qualitative} again: subjects fuse with each other or with the background, smear into ghosts, or leave the shot halfway through the clip.
\method keeps each subject whole from the first frame to the last while it still moves.

\begin{figure*}[t]
\centering
\includegraphics[width=0.9\textwidth]{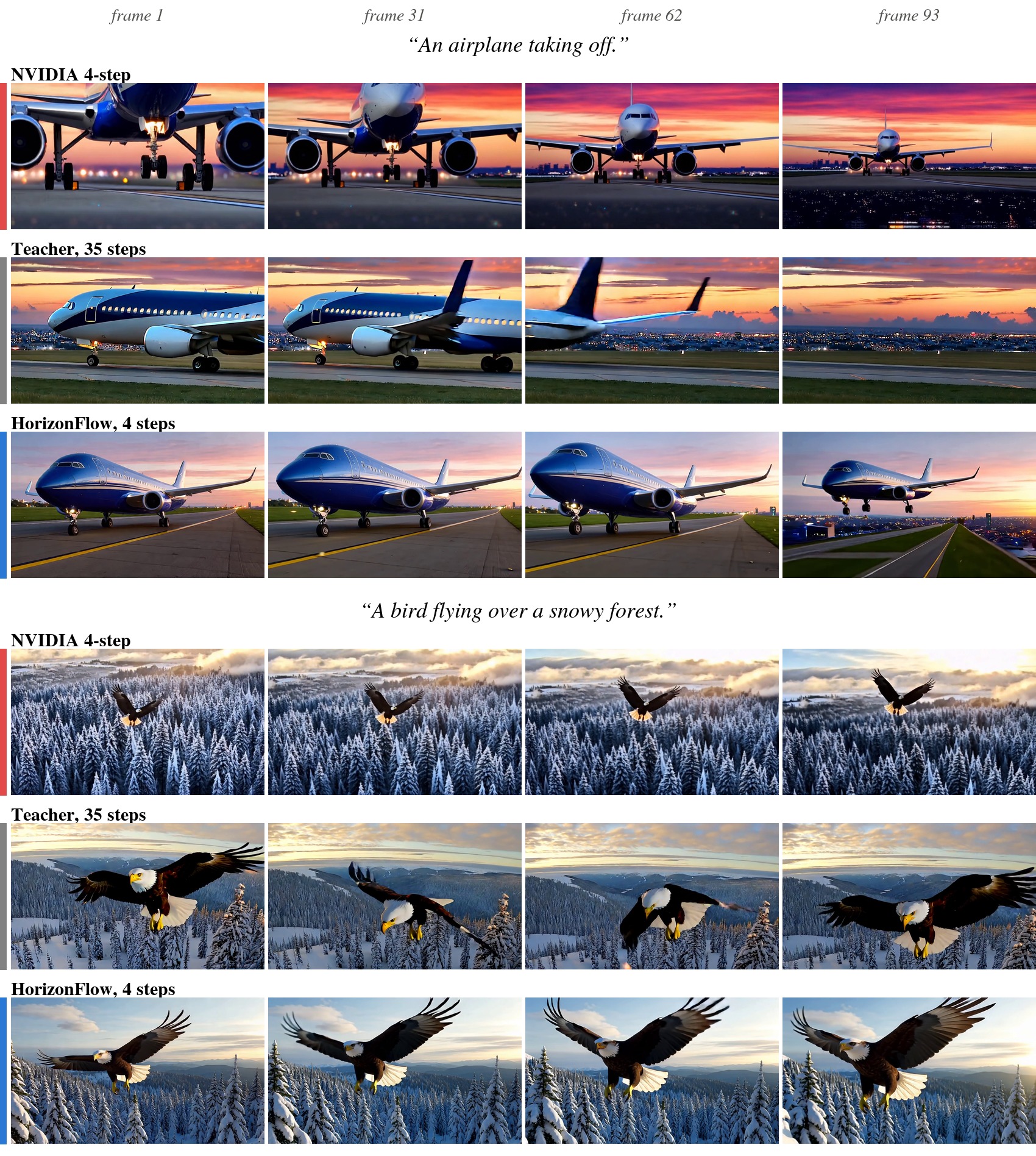}
\caption{\textbf{Cosmos-Predict2.5-2B: NVIDIA's 4-step model, the 35-step teacher, and \method.}
Same prompt and seed; frames 1, 31, 62 and 93 of 93.
NVIDIA's 4-step model keeps the airplane on the runway and the eagle hovering in place.
\method, distilled with the unchanged recipe and no real video, lifts the airplane off the runway and lets the eagle beat its wings over the forest.}
\label{fig:supp_cosmos}
\end{figure*}

\paragraph{Cosmos-Predict2.5-2B.}
\Cref{fig:supp_cosmos} compares the 4-step students on Cosmos with prompts from VBench's dynamic-degree set.
Where NVIDIA's distilled model stays nearly static, \method carries out the motion the prompt asks for.

\section{Failure cases}
\label{sec:supp_fail}
\begin{figure*}[p]
\centering
\includegraphics[width=\textwidth,height=0.84\textheight,keepaspectratio]{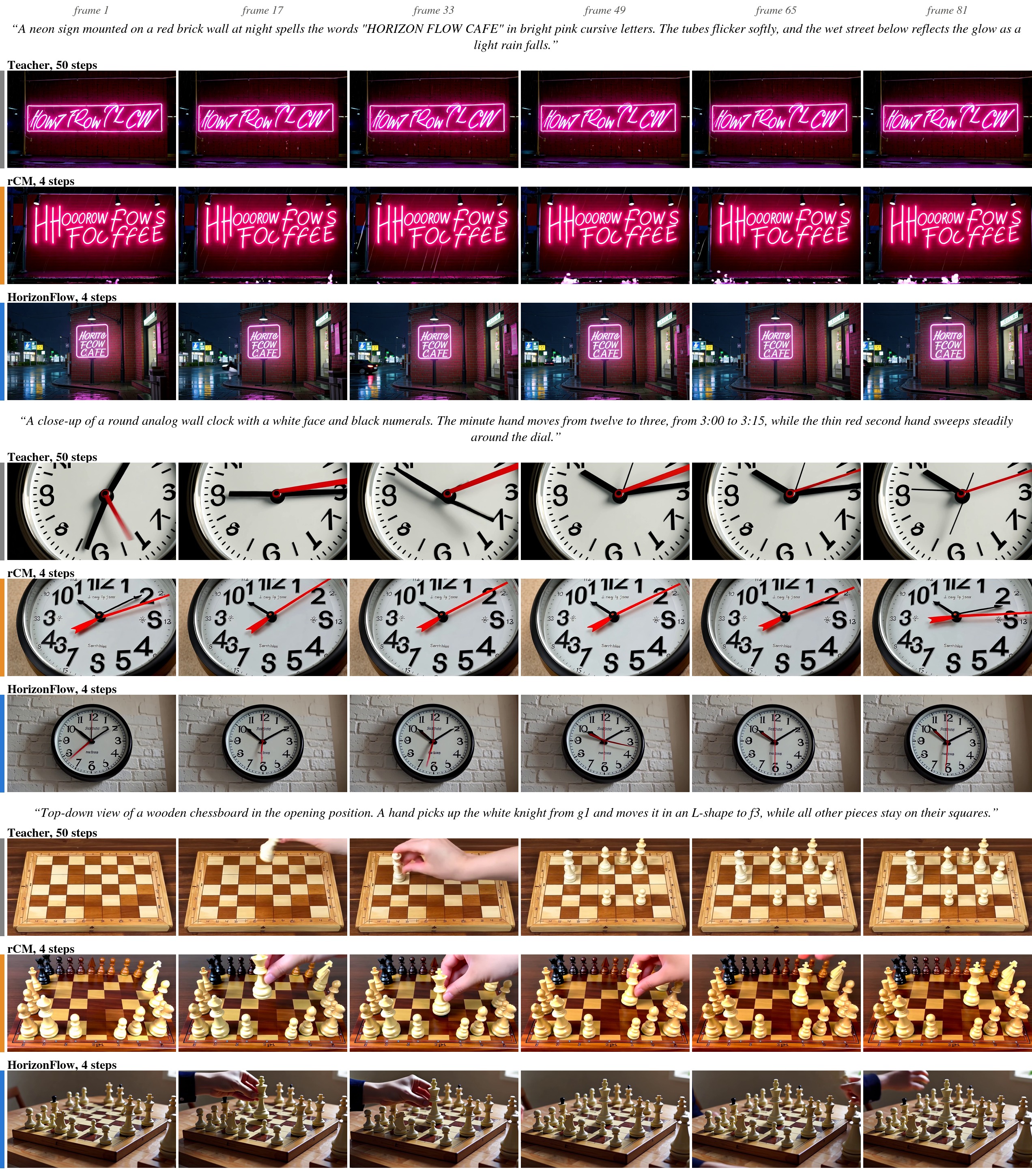}
\caption{\textbf{Failure cases beyond VBench on Wan2.1-T2V-1.3B.}
Same prompt and seed for the 50-step teacher, rCM and \method; frames 1, 17, 33, 49, 65 and 81 of 81.
All three models misspell the sign, and only \method spells ``CAFE'' correctly.
No model turns the clock from 3:00 to 3:15. The teacher and rCM also scramble the numerals, while \method lays them out in order.
No model sets up the opening position or moves the knight, and the teacher starts from an empty board on which white pieces appear.
\method keeps each scene stable even where the prompt asks for more than the teacher can do.}
\label{fig:supp_failure}
\end{figure*}

\Cref{fig:supp_failure} shows three prompts beyond the reach of the 50-step teacher: rendering a given text, showing an exact time, and following the rules of chess.
The teacher fails on all three, and so do rCM and \method, on both seeds we generated (42 and 43).
DMD trains the student to match the distribution of the teacher, so the student inherits what the teacher cannot do.
CE and SC supply the two signals that the DMD gradient lacks, one for even motion within latent chunks and one for keeping the committed layout.
Neither adds knowledge that the teacher lacks.
Even in these failures, \method keeps each scene stable.
Since the recipe carries over to another backbone (\cref*{tab:transfer}), a stronger teacher can be distilled the same way.

{
    \small
\begin{thebibliography}{36}
\providecommand{\natexlab}[1]{#1}
\providecommand{\url}[1]{\texttt{#1}}
\expandafter\ifx\csname urlstyle\endcsname\relax
  \providecommand{\doi}[1]{doi: #1}\else
  \providecommand{\doi}{doi: \begingroup \urlstyle{rm}\Url}\fi

\bibitem[Ali et~al.(2025)Ali, Bai, Bala, Balaji, Blakeman, Cai, Cao, Cao, Cha,
  Chao, et~al.]{cosmos25}
Arslan Ali, Junjie Bai, Maciej Bala, Yogesh Balaji, Aaron Blakeman, Tiffany
  Cai, Jiaxin Cao, Tianshi Cao, Elizabeth Cha, Yu-Wei Chao, et~al.
\newblock World simulation with video foundation models for physical ai.
\newblock \emph{arXiv preprint arXiv:2511.00062}, 2025.

\bibitem[Duan et~al.(2026)Duan, Xiang, Chen, Zhang, Liu, Huang, Zhang, Fan, and
  Li]{udmd}
Zixuan Duan, Xunzhi Xiang, Yabo Chen, Xin Zhang, Changhan Liu, Haibin Huang,
  Chi Zhang, Qi Fan, and Xuelong Li.
\newblock Uncertainty dmd: Restoring diversity in few-step autoregressive video
  distillation.
\newblock \emph{arXiv preprint arXiv:2609.11265}, 2026.

\bibitem[Fan et~al.(2026)Fan, Qiu, Wu, Wang, Lin, Ren, Lin, Gong, and
  Yang]{phaseddmd}
Xiangyu Fan, Zesong Qiu, Zhuguanyu Wu, Fanzhou Wang, Zhiqian Lin, Tianxiang
  Ren, Dahua Lin, Ruihao Gong, and Lei Yang.
\newblock Phased dmd: Few-step distribution matching distillation via score
  matching within subintervals.
\newblock In \emph{Proceedings of the IEEE/CVF Conference on Computer Vision
  and Pattern Recognition}, pages 41667--41676, 2026.

\bibitem[Gu et~al.(2026)Gu, Fang, Jiang, Mao, Han, Cai, and Shou]{anyflow}
Yuchao Gu, Guian Fang, Yuxin Jiang, Weijia Mao, Song Han, Han Cai, and
  Mike~Zheng Shou.
\newblock Anyflow: Any-step video diffusion model with on-policy flow map
  distillation.
\newblock In \emph{European Conference on Computer Vision}, pages 163--181.
  Springer, 2026.

\bibitem[Heusel et~al.(2017)Heusel, Ramsauer, Unterthiner, Nessler, and
  Hochreiter]{ttur}
Martin Heusel, Hubert Ramsauer, Thomas Unterthiner, Bernhard Nessler, and Sepp
  Hochreiter.
\newblock Gans trained by a two time-scale update rule converge to a local nash
  equilibrium.
\newblock \emph{Advances in neural information processing systems}, 30, 2017.

\bibitem[Ho and Salimans(2022)]{cfg}
Jonathan Ho and Tim Salimans.
\newblock Classifier-free diffusion guidance.
\newblock \emph{arXiv preprint arXiv:2207.12598}, 2022.

\bibitem[Ho et~al.(2022)Ho, Salimans, Gritsenko, Chan, Norouzi, and Fleet]{vdm}
Jonathan Ho, Tim Salimans, Alexey Gritsenko, William Chan, Mohammad Norouzi,
  and David Fleet.
\newblock Video diffusion models.
\newblock \emph{Advances in neural information processing systems},
  35:\penalty0 8633--8646, 2022.

\bibitem[Huang et~al.(2026)Huang, Li, He, Zhou, and Shechtman]{selfforcing}
Xun Huang, Zhengqi Li, Guande He, Mingyuan Zhou, and Eli Shechtman.
\newblock Self forcing: Bridging the train-test gap in autoregressive video
  diffusion.
\newblock \emph{Advances in Neural Information Processing Systems},
  38:\penalty0 167283--167308, 2026.

\bibitem[Huang et~al.(2024)Huang, He, Yu, Zhang, Si, Jiang, Zhang, Wu, Jin,
  Chanpaisit, et~al.]{vbench}
Ziqi Huang, Yinan He, Jiashuo Yu, Fan Zhang, Chenyang Si, Yuming Jiang, Yuanhan
  Zhang, Tianxing Wu, Qingyang Jin, Nattapol Chanpaisit, et~al.
\newblock Vbench: Comprehensive benchmark suite for video generative models.
\newblock In \emph{2024 IEEE/CVF Conference on Computer Vision and Pattern
  Recognition (CVPR)}, pages 21807--21818. IEEE, 2024.

\bibitem[Kong et~al.(2024)Kong, Tian, Zhang, Min, Dai, Zhou, Xiong, Li, Wu,
  Zhang, et~al.]{hunyuanvideo}
Weijie Kong, Qi Tian, Zijian Zhang, Rox Min, Zuozhuo Dai, Jin Zhou, Jiangfeng
  Xiong, Xin Li, Bo Wu, Jianwei Zhang, et~al.
\newblock Hunyuanvideo: A systematic framework for large video generative
  models.
\newblock \emph{arXiv preprint arXiv:2412.03603}, 2024.

\bibitem[Lin et~al.(2025)Lin, Xia, Ren, Yang, Xiao, and Jiang]{apt}
Shanchuan Lin, Xin Xia, Yuxi Ren, Ceyuan Yang, Xuefeng Xiao, and Lu Jiang.
\newblock Diffusion adversarial post-training for one-step video generation.
\newblock \emph{arXiv preprint arXiv:2501.08316}, 2025.

\bibitem[Lipman et~al.(2022)Lipman, Chen, Ben-Hamu, Nickel, and
  Le]{flowmatching}
Yaron Lipman, Ricky~TQ Chen, Heli Ben-Hamu, Maximilian Nickel, and Matt Le.
\newblock Flow matching for generative modeling.
\newblock \emph{arXiv preprint arXiv:2210.02747}, 2022.

\bibitem[Liu et~al.(2022)Liu, Gong, and Liu]{rectifiedflow}
Xingchao Liu, Chengyue Gong, and Qiang Liu.
\newblock Flow straight and fast: Learning to generate and transfer data with
  rectified flow.
\newblock \emph{arXiv preprint arXiv:2209.03003}, 2022.

\bibitem[Lu and Song(2025)]{scm}
Cheng Lu and Yang Song.
\newblock Simplifying, stabilizing and scaling continuous-time consistency
  models.
\newblock In \emph{International Conference on Learning Representations}, pages
  50611--50649, 2025.

\bibitem[Luo et~al.(2023)Luo, Tan, Huang, Li, and Zhao]{lcm}
Simian Luo, Yiqin Tan, Longbo Huang, Jian Li, and Hang Zhao.
\newblock Latent consistency models: Synthesizing high-resolution images with
  few-step inference.
\newblock \emph{arXiv preprint arXiv:2310.04378}, 2023.

\bibitem[Nie et~al.(2026)Nie, Berner, Ma, Liu, Xie, and Vahdat]{tmd}
Weili Nie, Julius Berner, Nanye Ma, Chao Liu, Saining Xie, and Arash Vahdat.
\newblock Transition matching distillation for fast video generation.
\newblock \emph{arXiv preprint arXiv:2601.09881}, 2026.

\bibitem[Peebles and Xie(2023)]{dit}
William Peebles and Saining Xie.
\newblock Scalable diffusion models with transformers.
\newblock In \emph{2023 IEEE/CVF International Conference on Computer Vision
  (ICCV)}, pages 4172--4182. IEEE, 2023.

\bibitem[Rombach et~al.(2022)Rombach, Blattmann, Lorenz, Esser, and Ommer]{ldm}
Robin Rombach, Andreas Blattmann, Dominik Lorenz, Patrick Esser, and Bj{\"o}rn
  Ommer.
\newblock High-resolution image synthesis with latent diffusion models.
\newblock In \emph{2022 IEEE/CVF conference on computer vision and pattern
  recognition (CVPR)}, pages 10674--10685. ieee, 2022.

\bibitem[Salimans and Ho(2022)]{progressive}
Tim Salimans and Jonathan Ho.
\newblock Progressive distillation for fast sampling of diffusion models.
\newblock \emph{arXiv preprint arXiv:2202.00512}, 2022.

\bibitem[Sauer et~al.(2024)Sauer, Lorenz, Blattmann, and Rombach]{add}
Axel Sauer, Dominik Lorenz, Andreas Blattmann, and Robin Rombach.
\newblock Adversarial diffusion distillation.
\newblock In \emph{European Conference on Computer Vision}, pages 87--103.
  Springer, 2024.

\bibitem[Song et~al.(2023)Song, Dhariwal, Chen, and Sutskever]{cm}
Yang Song, Prafulla Dhariwal, Mark Chen, and Ilya Sutskever.
\newblock Consistency models.
\newblock \emph{arXiv preprint arXiv:2303.01469}, 2023.

\bibitem[Wan et~al.(2025)Wan, Wang, Ai, Wen, Mao, Xie, Chen, Yu, Zhao, Yang,
  et~al.]{wan2025}
Team Wan, Ang Wang, Baole Ai, Bin Wen, Chaojie Mao, Chen-Wei Xie, Di Chen,
  Feiwu Yu, Haiming Zhao, Jianxiao Yang, et~al.
\newblock Wan: Open and advanced large-scale video generative models.
\newblock \emph{arXiv preprint arXiv:2503.20314}, 2025.

\bibitem[Wang et~al.(2023)Wang, Lu, Wang, Bao, Li, Su, and Zhu]{vsd}
Zhengyi Wang, Cheng Lu, Yikai Wang, Fan Bao, Chongxuan Li, Hang Su, and Jun
  Zhu.
\newblock Prolificdreamer: High-fidelity and diverse text-to-3d generation with
  variational score distillation.
\newblock \emph{Advances in neural information processing systems},
  36:\penalty0 8406--8441, 2023.

\bibitem[Wang et~al.(2026)Wang, Yuan, Wang, Yang, Zhang, Yuan, Huang, Liu,
  Wang, Yang, et~al.]{pdmd}
Zimo Wang, Junkun Yuan, Angtian Wang, Haotian Yang, Canyu Zhang, Siyuan Yuan,
  Xingchang Huang, Bo Liu, Yizhi Wang, Yiding Yang, et~al.
\newblock Pdmd: Projected distribution matching distillation for video
  diffusion models.
\newblock \emph{arXiv preprint arXiv:2609.35768}, 2026.

\bibitem[Wu et~al.(2026)Wu, Gong, Yong, Huang, Fan, Yang, Lin, and Liu]{sgmd}
Zhuguanyu Wu, Ruihao Gong, Yang Yong, Yushi Huang, Xiangyu Fan, Lei Yang, Dahua
  Lin, and Xianglong Liu.
\newblock Sgmd: Score gradient matching distillation for few-step video
  diffusion distillation.
\newblock \emph{arXiv preprint arXiv:2605.30116}, 2026.

\bibitem[Xu et~al.(2026)Xu, Huang, Lu, Zhong, Fan, Huang, and Liu]{dymd}
Haojun Xu, Jie Huang, Xin Lu, Mingchen Zhong, Zihao Fan, Linjiang Huang, and Si
  Liu.
\newblock Dymd: Preserving interaction dynamics through distribution matching
  distillation in few-step video world models.
\newblock \emph{arXiv preprint arXiv:2609.31349}, 2026.

\bibitem[Yang et~al.(2025)Yang, Teng, Zheng, Ding, Huang, Xu, Yang, Hong,
  Zhang, Feng, et~al.]{cogvideox}
Zhuoyi Yang, Jiayan Teng, Wendi Zheng, Ming Ding, Shiyu Huang, Jiazheng Xu,
  Yuanming Yang, Wenyi Hong, Xiaohan Zhang, Guanyu Feng, et~al.
\newblock Cogvideox: Text-to-video diffusion models with an expert transformer.
\newblock In \emph{International Conference on Learning Representations}, pages
  83048--83077, 2025.

\bibitem[Yin et~al.(2024{\natexlab{a}})Yin, Gharbi, Park, Zhang, Shechtman,
  Durand, and Freeman]{dmd2}
Tianwei Yin, Micha{\"e}l Gharbi, Taesung Park, Richard Zhang, Eli Shechtman,
  Fredo Durand, and William~T Freeman.
\newblock Improved distribution matching distillation for fast image synthesis.
\newblock \emph{Advances in neural information processing systems},
  37:\penalty0 47455--47487, 2024{\natexlab{a}}.

\bibitem[Yin et~al.(2024{\natexlab{b}})Yin, Gharbi, Zhang, Shechtman, Durand,
  Freeman, and Park]{dmd}
Tianwei Yin, Micha{\"e}l Gharbi, Richard Zhang, Eli Shechtman, Fredo Durand,
  William~T Freeman, and Taesung Park.
\newblock One-step diffusion with distribution matching distillation.
\newblock In \emph{2024 IEEE/CVF Conference on Computer Vision and Pattern
  Recognition (CVPR)}, pages 6613--6623. IEEE, 2024{\natexlab{b}}.

\bibitem[Yin et~al.(2025)Yin, Zhang, Zhang, Freeman, Durand, Shechtman, and
  Huang]{causvid}
Tianwei Yin, Qiang Zhang, Richard Zhang, William~T Freeman, Fredo Durand, Eli
  Shechtman, and Xun Huang.
\newblock From slow bidirectional to fast autoregressive video diffusion
  models.
\newblock In \emph{2025 IEEE/CVF Conference on Computer Vision and Pattern
  Recognition (CVPR)}, pages 22963--22974. IEEE, 2025.

\bibitem[You et~al.(2026)You, Li, Li, Mu, Chen, and Jiang]{adv}
Yuyang You, Yongzhi Li, Jiahui Li, Yadong Mu, Quan Chen, and Peng Jiang.
\newblock Adaptive video distillation: Mitigating oversaturation and temporal
  collapse in few-step generation.
\newblock \emph{arXiv preprint arXiv:2603.21864}, 2026.

\bibitem[Yu et~al.(2026)Yu, Yuan, Yang, Qian, Wang, Wang, Yang, Liu, Li, Wang,
  et~al.]{dmad}
Zhengming Yu, Junkun Yuan, Haotian Yang, Gordon~Guocheng Qian, Yizhi Wang,
  Angtian Wang, Yiding Yang, Bo Liu, Xin Li, Wenping Wang, et~al.
\newblock Dmad: Distribution matching as adversarial distillation for fast
  visual generation.
\newblock \emph{arXiv preprint arXiv:2610.02188}, 2026.

\bibitem[Zhang et~al.(2025)Zhang, Chen, Huang, Lin, Liu, Stoica, Xing, and
  Zhang]{vsa}
Peiyuan Zhang, Yongqi Chen, Haofeng Huang, Will Lin, Zhengzhong Liu, Ion
  Stoica, Eric Xing, and Hao Zhang.
\newblock Vsa: Faster video diffusion with trainable sparse attention, 2025.

\bibitem[Zhang et~al.(2026)Zhang, Cheng, Wang, Li, Zhang, Li, Zha, Zhang, Zhao,
  and Jia]{codmd}
Wenhu Zhang, Kun Cheng, Changyuan Wang, Shiyao Li, Yuechen Zhang, Wenbo Li,
  Jiajun Zha, Jingyi Zhang, Kang Zhao, and Jiaya Jia.
\newblock Codmd: Copula-aware distribution matching distillation for fast video
  generation.
\newblock \emph{arXiv preprint arXiv:2606.21982}, 2026.

\bibitem[Zheng et~al.(2026)Zheng, Wang, Ma, Chen, Zhang, Balaji, Chen, Liu,
  Zhu, and Zhang]{rcm}
Kaiwen Zheng, Yuji Wang, Qianli Ma, Huayu Chen, Jintao Zhang, Yogesh Balaji,
  Jianfei Chen, Ming-Yu Liu, Jun Zhu, and Qinsheng Zhang.
\newblock Large scale diffusion distillation via score-regularized
  continuous-time consistency.
\newblock In \emph{International Conference on Learning Representations}, pages
  2582--2603, 2026.

\bibitem[Zhou et~al.(2024)Zhou, Zheng, Wang, Yin, and Huang]{sid}
Mingyuan Zhou, Huangjie Zheng, Zhendong Wang, Mingzhang Yin, and Hai Huang.
\newblock Score identity distillation: Exponentially fast distillation of
  pretrained diffusion models for one-step generation.
\newblock In \emph{Forty-first International Conference on Machine Learning},
  2024.

\end{thebibliography}
}
\end{document}